\documentclass[11pt]{article}

\usepackage[preprint]{acl}

\usepackage{times}
\usepackage{latexsym}

\usepackage[T1]{fontenc}

\usepackage[utf8]{inputenc}

\usepackage{microtype}

\usepackage{inconsolata}

\usepackage{graphicx}

\usepackage{url}            
\usepackage{booktabs}       
\usepackage{amsfonts}       
\usepackage{nicefrac}       
\usepackage{amsmath}  
\usepackage[cjk]{kotex} 
\usepackage{mdframed}
\usepackage{multirow}
\usepackage{subcaption}
\usepackage[table,xcdraw]{xcolor}
\usepackage{makecell}
\usepackage{pifont}
\usepackage{CJKutf8}
\usepackage{tabularx}
\usepackage{wrapfig}
\usepackage{enumitem}
\usepackage{xspace}
\usepackage[most]{tcolorbox}
\usepackage{hyperref}
\usepackage{ulem}
\newcommand\highlight[1][yellow]{%
  \bgroup 
  \markoverwith{\textcolor{#1}{\vrule width.15em height.85em depth.2em}}%
  \ULon 
}

\newcommand{\hlinst}[1]{\highlight[gray!30]{#1}}

\newcommand{\cmark}{\text{\ding{51}}}
\newcommand{\xmark}{\text{\ding{55}}}
\def\eg{\emph{e.g}.,\xspace}
\def\ie{\emph{i.e}.,\xspace}
\newcolumntype{P}[1]{>{\centering\arraybackslash}p{#1}}

\definecolor{redhighlight}{rgb}{1, 0.8, 0.8} 
\newcommand{\inc}[1]{{\color{teal}\textbf{(+#1)}}}
\newcommand{\dec}[1]{{\color{red}(#1)}}

\newif\ifcommentsoff

\title{OLA: Output Language Alignment in Code-Switched LLM Interactions}

\author{Juhyun Oh \hspace{1em} Haneul Yoo\thanks{Equal contribution.} \hspace{1em} Faiz Ghifari Haznitrama$^*$  \hspace{1em} Alice Oh \\
        KAIST \\
        \texttt{\href{mailto:411juhyun@kaist.ac.kr}{\{411juhyun}, 
            \href{mailto:haneul.yoo@kaist.ac.kr}{haneul.yoo}, 
            \href{mailto:haznitrama@kaist.ac.kr}{haznitrama\}@kaist.ac.kr}, 
            \href{mailto:alice.oh@kaist.edu}{alice.oh@kaist.edu}
        }}
\begin{document}
\maketitle
\begin{abstract}
Code-switching, alternating between languages within a conversation, is natural for multilingual users, yet poses fundamental challenges for large language models (LLMs). When a user code-switches in their prompt to an LLM, they typically do not specify the expected language of the LLM response, and thus LLMs must infer the output language from contextual and pragmatic cues. We find that current LLMs systematically fail to align with this expectation, responding in undesired languages even when cues are clear to humans.
We introduce \textsc{OLA}, a benchmark to evaluate LLMs' \textbf{O}utput \textbf{L}anguage \textbf{A}lignment in code-switched interactions. 
\textsc{OLA} focuses on Korean--English code-switching and spans simple intra-sentential mixing to instruction–content mismatches. Even frontier models frequently misinterpret implicit language expectation, exhibiting a bias toward non-English responses. We further show this bias generalizes beyond Korean to Chinese and Indonesian pairs. Models also show instability through mid-response switching and language intrusions. Chain-of-Thought prompting fails to resolve these errors, indicating weak pragmatic reasoning about output language.
However, Code-Switching Aware DPO with minimal data (${\sim}$1K examples) substantially reduces misalignment, suggesting these failures stem from insufficient alignment rather than fundamental limitations. Our results highlight the need to align multilingual LLMs with users' implicit expectations in real-world code-switched interactions.\footnote{OLA is available at \url{https://github.com/juhyunohh/OLA}}
\end{abstract}

\section{Introduction}
\begin{figure}[t]
  \centering
  \includegraphics[width=\linewidth]{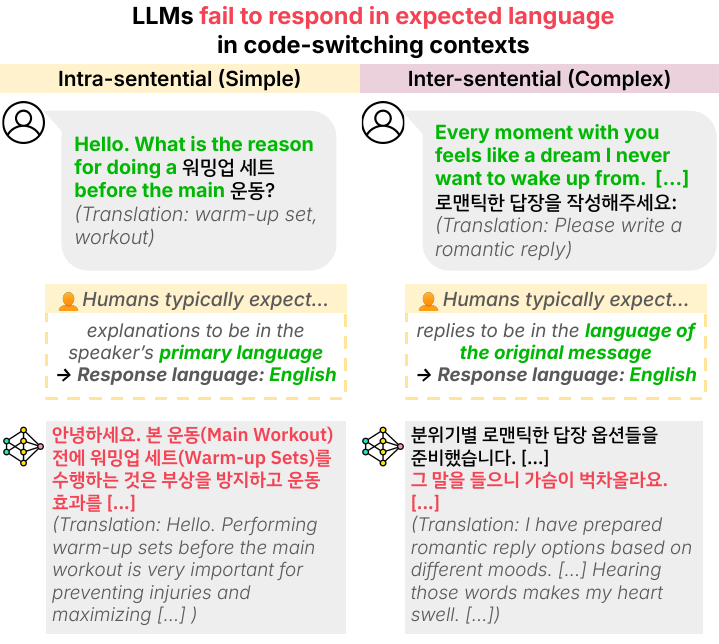}
  \caption{\textbf{Illustrative failures under intra- and inter-sentential code-switching.} The model understands the task content but misses the user's implicit expectation about the output language, causing avoidable friction (intra) and mis-execution of the request with obvious language expectations (inter).}
  \label{fig:teaser}
\end{figure}

\normalem
For a large population of multilingual users, interacting with large language models (LLMs) naturally involves \emph{code-switching}---the interleaving of multiple languages within a single conversational context~\cite{auer1998code}. A common scenario is to provide content in one language (\eg an English draft) while issuing an instruction in another (\eg requesting edits in Korean). In such cases, the expected output language is typically left implicit, governed by pragmatic and contextual cues. While humans readily infer this expectation, current LLMs often fail to do so, producing responses in unexpected languages and forcing users to over-specify what would otherwise be obvious. Figure~\ref{fig:teaser} illustrates representative misalignment cases.

Prior work has examined the problem of models responding in undesired languages mainly through monolingual benchmarks~\cite{marchisio-etal-2024-understanding, nie-etal-2025-mechanistic}. However, these settings overlook a more challenging and practical scenario: code-switched inputs, where multiple languages coexist and the output language must be inferred from context.
Existing studies on code-switching have focused on generating linguistically plausible mixed text~\cite{xie2025switchlingua, kuwanto2024linguisticstheorymeetsllm} or task performance degradation~\cite{zhang-etal-2023-multilingual}, leaving open whether models can \emph{respond in the language users implicitly expect}.

In this work, we address this gap by systematically studying \emph{output language misalignment in code-switched interactions}: a model's failure to infer and follow the user's expected output language, despite clear contextual cues. 
We introduce \textsc{OLA}(Output Language Alignment) benchmark that evaluates this ability across two common patterns of code-switching: intra-sentential mixing and instruction--content mismatches.

Our analysis reveals three key findings. 
First, output language misalignment in code-switched prompts is systematic: models exhibit strong asymmetries, disproportionately defaulting to non-English responses even when English is the primary language, a pattern that generalizes to other language pairs (English--Chinese, English--Indonesian). 
Second, failures extend beyond binary language choice, manifesting as mid-response language intrusions. 
Third, these failures cannot be resolved through inference-time reasoning alone: Chain-of-Thought prompting often degrades performance, indicating models' lack of internal criteria to infer implicit language expectation.

We show that these failures can be substantially mitigated through targeted alignment. Code-Switching Aware DPO (CS-DPO), trained on only ${\sim}$1K simple intra-sentential examples, significantly improves output language alignment and generalizes to more complex settings and unseen language pairs. This indicates that misalignment primarily stems from insufficient alignment rather than fundamental modeling limits.

In summary, we make three contributions: 
(1) we identify output language misalignment in code-switched interactions as a critical blind spot in multilingual LLMs and introduce a systematic benchmark; 
(2) we show that models rely on surface heuristics, leading to systematic misalignment under implicit language constraints; 
(3) we demonstrate that targeted preference alignment with minimal data substantially improves performance and generalizes across language pairs and complexity levels.
More broadly, our work highlights the need for multilingual LLMs to move beyond isolated monolingual competence and align with implicit user expectations in code-switched interactions.



\section{Background and Related Work}
\normalem
\paragraph{Language Confusion and Output Language Misalignment.}
Prior work has shown that multilingual LMs may generate output in undesired languages, broadly referred to as \emph{language confusion}.
In machine translation, this phenomenon is studied as \emph{off-target translation}, where a model fails to translate into the explicitly specified target language~\citep{zhang-etal-2020-improving, wu-etal-2021-language, chen-etal-2023-target}.
More recently, \citet{marchisio-etal-2024-understanding} show that LLMs may violate both explicit language instructions (\eg ``Respond in Hindi'') and implicit monolingual expectations (\ie output to Hindi input should be Hindi), producing undesired languages at the response, sentence, or word level.

While these studies establish that language control remains imperfect in multilingual LLMs, they primarily examine failures of \emph{instruction following} or \emph{language consistency}.
In contrast, our work focuses on a distinct and practically important source of error: output language \emph{misalignment} that arises when the response language is not explicitly specified and must instead be inferred from contextual and pragmatic cues.
We focus in particular on such misalignment in code-switched interactions, where multiple languages coexist in the input.

\begin{table*}[t] 
\small
\begin{tabularx}{\linewidth}{lXP{0.13\linewidth}P{0.15\linewidth}}
\toprule
Setting & Prompt & Type & Expected Lang. \\ \midrule
Simple & 
\textcolor{magenta}{Hello. What is the reason for doing a} 워밍업 세트 \textcolor{magenta}{before the main} 운동?

(Translation: \textit{Hello. What is the reason for doing warm-up sets before the main exercise?}) & 
EN Matrix \newline-- KO Embed & 
\textcolor{magenta}{English} \newline (Matrix Lang.) \\ \midrule
Complex & 
\textcolor{magenta}{To eliminate sample bias, we can use different strategies like random sampling techniques, increasing sample size, employing stratified sampling to ensure representation from different subgroups,
[\ldots]}  
\newline \hlinst{이 문맥은 AI에 의해 작성되었나요?}

(Translation: \textit{Was this passage written by an AI?}) & 
KO Instr. \newline-- EN Content & 
{Korean} \newline (Instruction Lang.) \\ \bottomrule
\end{tabularx}
\caption{\textbf{Representative examples from OLA Benchmark.} Text colors indicate language (\textcolor{magenta}{English in magenta}, Korean in black). In the Complex setting, \hlinst{highlighted text denotes the instruction segment}. The expected output language is determined by the matrix language in Simple settings and by task context in Complex settings. Monolingual source prompts for examples shown here are drawn from WildChat-1M~\cite{zhao2024wildchat}.}
\label{tab:dataset}
\end{table*}

\paragraph{Code-switching in Human--LLM Interactions.}
Code-switching, the use of multiple languages within a single conversational context, is a natural and pervasive behavior among multilingual speakers~\citep{winata-etal-2023-decades}.
Linguistically, code-switching can occur at multiple levels, including intra-word switching, tag-switching, intra-sentential switching, and inter-sentential switching.
In human--AI interaction, multilingual users frequently mix languages and expect models to handle such inputs appropriately~\citep{bawa2020do, choi2023toward}.

Empirical evidence suggests that such interactions overwhelmingly rely on implicit language expectation.
Our analysis of 335 Korean--English code-switched conversations from WildChat-4.8M~\citep{zhao2024wildchat} shows that only 45 (${\sim}$13\%) explicitly specify the desired output language, many of which correspond to translation requests.
In the remaining cases, users rely on contextual and pragmatic cues to signal their expectations.
Qualitative inspection further reveals two recurring interaction patterns illustrated in Figure~\ref{fig:teaser}:
(i) intra-sentential code-switching, where one language provides the grammatical frame while elements from another are embedded, and
(ii) instruction--content mismatches, where users issue instructions in one language while providing content in another.

Despite their prevalence, these interaction patterns remain underexplored in prior work.
Most studies on code-switching in LLMs focus on generating linguistically plausible mixed text~\citep{xie2025switchlingua, kuwanto2024linguisticstheorymeetsllm} or measuring task performance degradation under mixed-language inputs~\citep{zhang-etal-2023-multilingual}.
Other work analyzes language mixing in intermediate reasoning processes and its potential benefits for problem solving~\citep{li-etal-2025-impact, wang-etal-2025-language-mixing}.
In contrast, we focus on a distinct question grounded in real interactional settings: whether models can correctly infer and maintain the expected \emph{output language} when that expectation is implicit in a code-switched prompt.



\label{sec:background}

\section{OLA Benchmark}
\normalem


We introduce \textsc{OLA}, a benchmark for evaluating whether models correctly generate output in the expected language under code-switched prompts. \textsc{OLA} focuses on Korean--English code-switching and consists of two settings: Simple and Complex (Table~\ref{tab:dataset}).


The \textbf{Simple setting} focuses on intra-sentential code-switching, where the expected output language is the \emph{matrix language}---the language providing the core grammatical structure into which elements from another language are embedded~\citep{myers1997duelling}. For example, in ``What is the reason for doing a 워밍업 세트?'', English is the matrix language despite Korean noun phrases being embedded.
The \textbf{Complex setting} involves inter-sentential code-switching, where instruction and content languages differ, and the correct output language must be inferred from task semantics. This setting reflects a pattern that often emerges in human–AI interaction but has received limited attention in prior works, which largely focus on intra-sentential switching.

\subsection{Simple Setting}

\paragraph{Data Sources and Generation.}
We collect 299 initial prompts from the monolingual Language Confusion Benchmark~\citep{marchisio-etal-2024-understanding} (199 EN) and WildChat 1M~\citep{zhao2024wildchat} (100 KO), excluding queries that explicitly request translation or specify output language. To ensure consistent expectations, we convert text generation tasks (\eg ``Write an essay about...'') into question-answering format, as pilot annotations revealed that annotators are  sensitive to the language in which target objects are specified (\eg ``essay'' vs. ``에세이''), leading to low agreement.

Following \citet{kim2025codeswitched}, we generate code-switched queries by prompting an LLM with parallel sentences. We adopt noun phrase insertion as the code-switching strategy---the most common pattern among non-balanced multilinguals~\citep{lipski2014spanish, halpin2021code, zhong2024mandarin}---while strictly preserving matrix-language syntax (SVO for English, SOV for Korean)
All examples are synthesized using GPT-4o~\cite{openai2024gpt4ocard} with few-shot prompts (see Appendix~\ref{app:gen_OLA_prompts}).

\paragraph{Human Verification.}
Six native speakers (3 English, 3 Korean) who can understand and use the embedded language at a basic level, annotate the expected output language and rate the severity of language misalignment as \texttt{Trivial} (no retry needed), \texttt{Uncomfortable} (would prompt explicit language specification in subsequent interactions), or \texttt{Critical} (would reject and re-generate the response with an explicit language instruction). 
Retaining samples with agreement from at least two annotators yields 279 KO Matrix and 258 EN Matrix prompts. Over 92\% of all prompts are rated \texttt{Uncomfortable} or \texttt{Critical} for language mismatches, confirming strong user sensitivity.





\subsection{Complex Setting} 
\paragraph{Data Sources and Generation.} 
Based on an analysis of real-world code-switched interactions exhibiting instruction--content mismatches (Section~\ref{sec:background}), we identify two task categories with distinct language expectations:
\begin{enumerate}[leftmargin=*, topsep=0pt,itemsep=2pt,parsep=0pt]
    \item \textbf{Response in Instruction Language}: Content understanding tasks (\eg clarification, question-answering, summarizing) where users prefer responses in their more comfortable instruction language.
    \item \textbf{Response in Content Language}: Content manipulation tasks (\eg editing, revision, continuation) where outputs must maintain the original content's language.
\end{enumerate}
We curate 60 representative instruction templates (30 per category) from WildChat 1M~\citep{zhao2024wildchat} (see Appendix~\ref{sec:representative_template_queries}). For each template, we generate four content variations using GPT-4o and instantiate each with the instruction placed both before and after the content.

\paragraph{Human Verification.}
Using the same protocol as the Simple setting, we retain templates with robust agreement on expected output language. 
This yields 57 KO Instruction--EN Content templates (570 samples) and 54 EN Instruction--KO Content templates (540 samples). Over 90\% of templates are rated \texttt{Uncomfortable} or \texttt{Critical} by at least two annotators when output language mismatches the expected language.




\section{Experimental Setup}
\normalem

\paragraph{Models.}
We evaluate five multilingual LLMs: \textsc{Gemini 3 Pro Preview}~\cite{google2025gemini3promodelcard}, \textsc{GPT-5.1}~\cite{openai2025gpt5systemcard}, \textsc{Qwen 2.5 Instruct}~\cite{qwen2025qwen25}, \textsc{Exaone }4~\cite{research2025exaone4}, and \textsc{OLMo 3.1 Instruct-DPO}~\cite{olmo2025olmo}. 
All open-weight models are 32B-parameter variants.
The model cards and details are described in Appendix~\ref{app:main_experimental_setting}.

\paragraph{Metric.}
\begin{figure*}[tb!]
  \centering
  \begin{subfigure}[t]{0.35\linewidth}
    \centering
    \includegraphics[width=\linewidth]{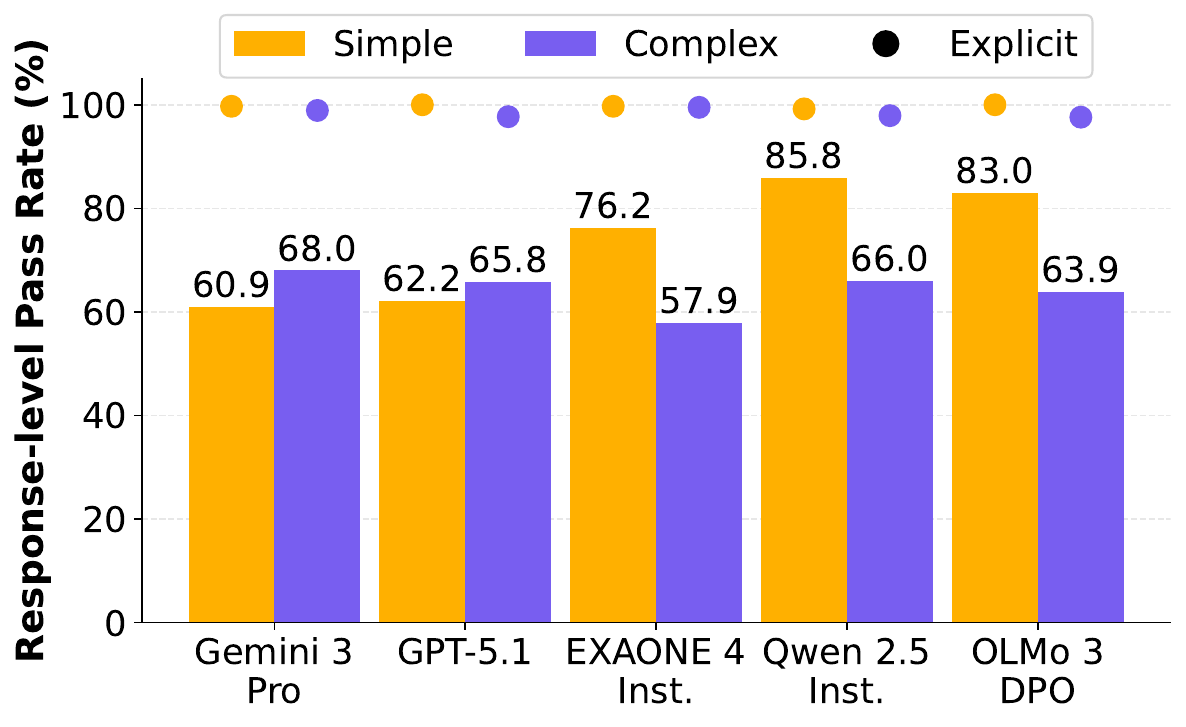}
    \caption{Overall result}
    \label{fig:main_simple_complex}
  \end{subfigure}
  \hfill
  \begin{subfigure}[t]{0.64\linewidth}
    \centering
    \includegraphics[width=\linewidth]{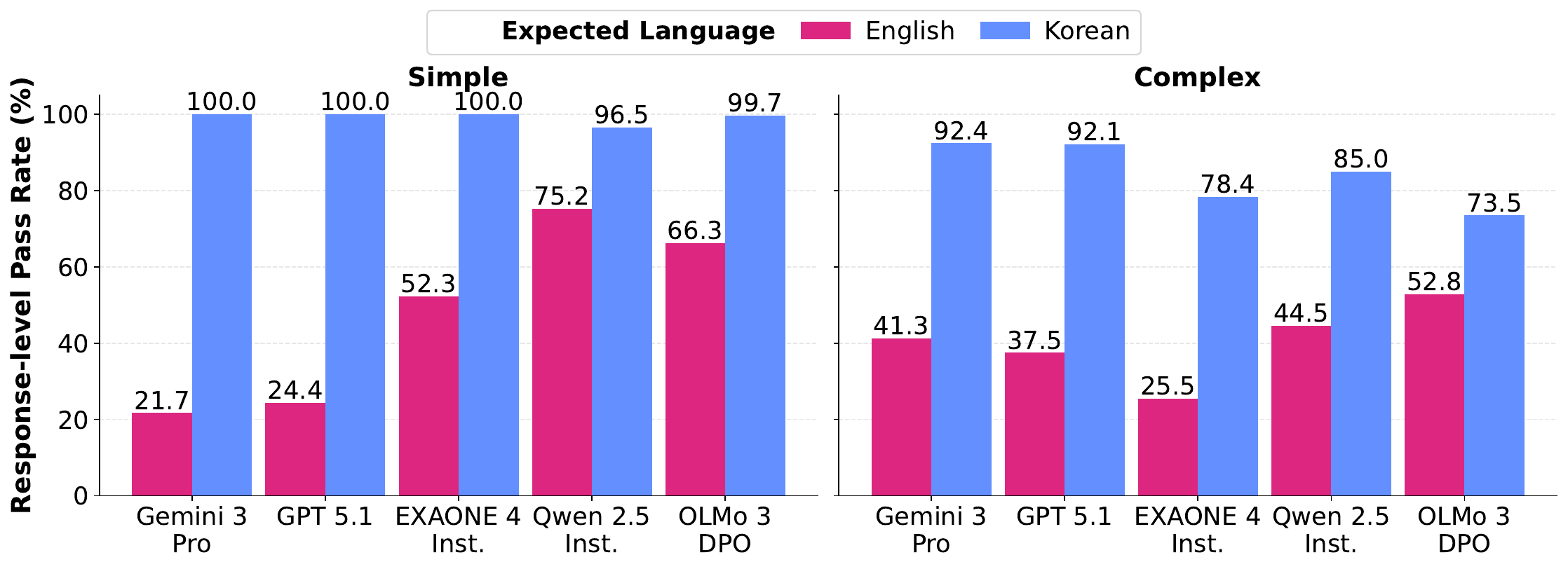}
    \caption{Detailed results by expected languages}
    \label{fig:main_by_language}
  \end{subfigure}
  \caption{\textbf{Response-level pass rates (\%) on the OLA benchmark.} (a) Overall performance in Simple and Complex settings. (b) Breakdown by expected output language.}
  \vspace{-1mm}
\end{figure*}

We evaluate model performance using \textit{Response-level Pass Rate}, which measures whether a response as a whole is generated in the expected language.
Following~\citet{marchisio-etal-2024-understanding}, we use fastText~\citep{joulin2016bag} for language identification with sentence-level majority voting: responses are split into sentences, each classified independently, and the primary language is determined by majority vote. This approach is robust to minor sentence-level code-switching.
A response is considered correct if its detected primary language matches the expected output language. We intentionally apply lenient evaluation: minor word- or line-level code-switching (\eg retained named entities or technical terms from the prompt) does not count as an error, as such behavior can reflect appropriate handling of code-switched inputs.

For Complex settings where instruction and content languages differ, meta-responses (\eg ``\textit{Here is the revised draft:}'') can confound response-level identification, as they may appear in the instruction language while the actual task content follows the content language.
We address this with \textit{Decompose-and-Verify} procedure: using GPT-4o\footnote{Qwen 2.5 Instruct yielded similar results.}, we segment responses into ``Meta-Response'' and ``Task-Content'', applying language identification only to the latter. This is used only when the expected language corresponds to the task content. Manual inspection of 652 GPT-4o-evaluated responses found only 5 errors (99.23\% accuracy).

\section{Main Results}
\normalem
\subsection{Systematic and Asymmetric Language Misalignment}
\label{sec:main_overall}
\paragraph{Overall Performance and Task Difficulty.}
Figure \ref{fig:main_simple_complex} shows that aligning with user expectations about output language poses substantial challenges for all evaluated models. Even in the Simple setting, where the matrix language directly signals the expected language, pass rates remain far from perfect (60.9--85.8\%). Performance deteriorates further in the Complex setting, where models must infer expected output language from task semantics(57.9--68.0\%).

To determine whether these failures stem from a lack of language proficiency or a failure in expectation inference, we evaluate an oracle condition with explicit language instructions (\eg ``\emph{Respond in 영어}''). All models achieve near-perfect pass rates under explicit specification (circle markers), confirming they possess adequate multilingual generation capabilities. This gap between explicit and implicit performance demonstrates that the bottleneck lies in inferring the user's expected language, not in generating responses in that language.

\paragraph{Asymmetric Bias Toward Non-English Responses.}
A closer analysis reveals that these failures are not randomly distributed but exhibit systematic asymmetry: models consistently favor generating non-English outputs. As shown in Figure \ref{fig:main_by_language}, pass rates drop sharply when English is the expected output in both Simple and Complex settings. This holds even when English is the matrix language (Simple) and when the instruction is given in English (Complex), suggesting models default to the non-English script present in the context. This asymmetry may reflect training data imbalance, where code-switched examples disproportionately involve non-English responses.

To test whether this bias generalizes beyond KO--EN code-switching, we extend our evaluation to Chinese--English and Indonesian--English Simple settings (see Appendix~\ref{app:multilingual} for details on dataset construction).
Across both language pairs, we observe the same qualitative pattern: when English is the expected output language (EN Matrix), models frequently generate responses in the non-English language instead. For instance, \textsc{GPT-5.1} achieves only 8.8\% pass rate in the {English Matrix--Chinese Embedded} configuration, despite near-perfect performance in the reverse configuration (see Table~\ref{tab:multilingual_failure} for full results).
Comparable degradation is observed in Indonesian--English settings as well, even though Indonesian shares the Latin script with English. These results demonstrate that the ``non-English default'' bias is not specific to Korean or to non-Latin scripts, but reflects a general failure mode in resolving implicit language expectation under code-switched prompts.

\begin{figure}[htb!]
  \centering
  \includegraphics[width=\linewidth]{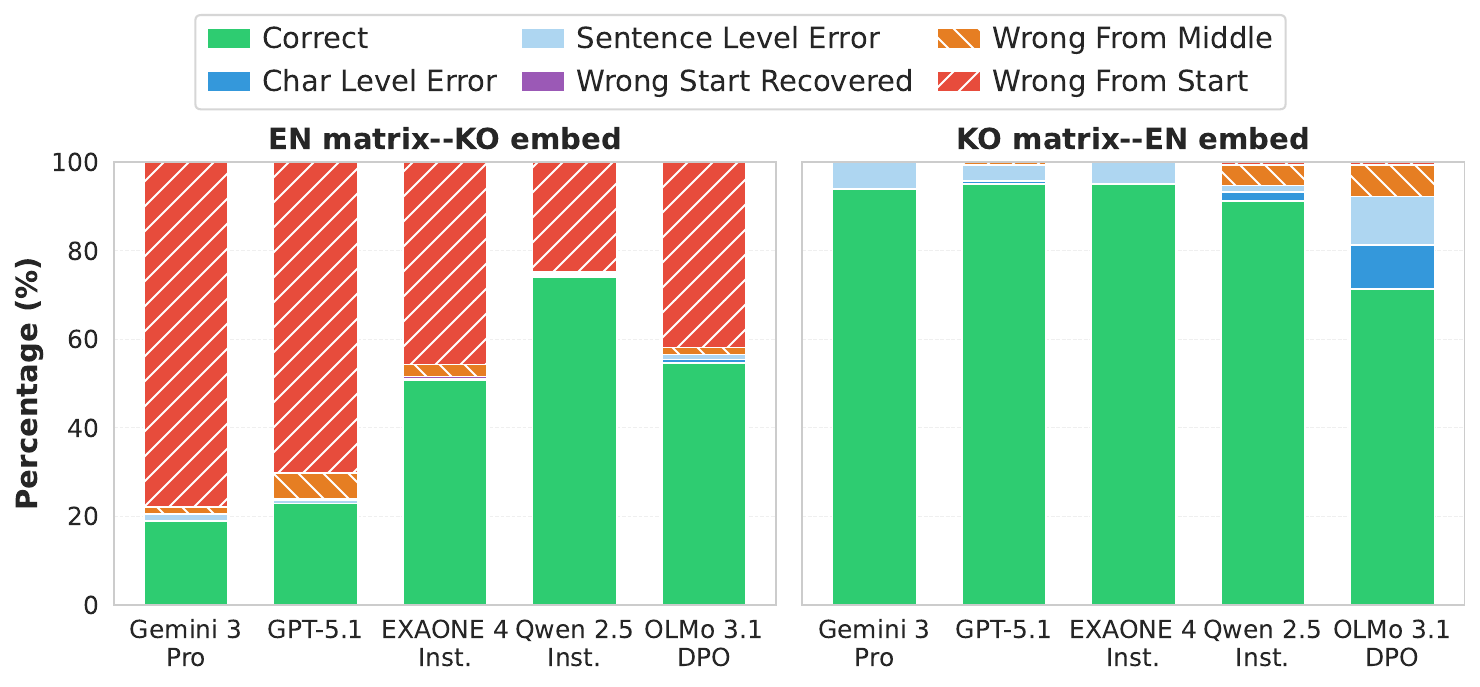}
  \caption{\textbf{Distribution of output language outcomes and error types in the Simple code-switched setting.} Hatched segments indicate output-level misalignment; other patterns show intrusion errors within otherwise correct outputs.}
  \label{fig:failure_pattern_breakdown}
  \vspace{-2mm}
\end{figure}

\subsection{Failure Patterns and Error Type Distribution}
\label{sec:failure_pattern}
To better understand output-level failures, we categorize incorrect responses in the Simple setting by when and how language deviations occur. Figure~\ref{fig:failure_pattern_breakdown} shows error-type distributions across models.
We distinguish between \textit{output-level errors} (hatched segments), where the dominant language of the output is in the wrong language, and \textit{intrusion errors}, where tokens from unrelated third languages (\eg Japanese, Russian) spuriously appear within otherwise correct outputs. We exclude English, Korean, and Chinese from intrusion analysis: English and Korean may legitimately reference query content, while Chinese overlaps with Sino-Korean orthography.

When English is the matrix language (EN matrix--KO embed), errors are dominated by \textit{wrong-from-start} failures: models respond in incorrect output language from the beginning and rarely recover mid-generation (near-zero \textit{wrong-start-recovered} cases). In contrast, when Korean is the matrix language (KO matrix--EN embed), models achieve high output-level accuracy but exhibit non-negligible third-language intrusions at the sentence (avg:5.37\%) and character levels (avg:2.54\%), which constitute critical usability failures even when the nominal output language is correct. 
Character-level intrusions often appear as spurious script mixing within tokens (\eg 아놀드 슈워\textit{цeneg}거).
Notably, intrusions are substantially more frequent in alignment-tuned models than in base models, particularly for DPO- and RL-based variants (Table~\ref{tab:char_level_intrusions}). For instance, \textsc{OLMo 3.1 DPO} shows a relatively high rate of character-level intrusions, suggesting that alignment fine-tuning can introduce subword-level instability.




In the Complex setting, we additionally observe qualitative degradation: even when models correctly respond in Korean, they frequently rely on awkward transliterations or overly literal translations rather than natural expressions or established loanwords, consistent with translationese effects~\citep{zhao2024how, etxaniz-etal-2024-multilingual, zhong-etal-2025-language, bafna2025translation, schut2025multilingual}.
Detailed error definitions and illustrative examples are provided in Table~\ref{tab:failure_sample} in the Appendix.


\section{Analysis}
\normalem
In this section, we analyze how current LLMs select an output language in code-switched contexts.

\subsection{Surface Cues Dominate Output Language Selection}
Our results indicate that when the expected output language is left implicit, models do not robustly infer it from linguistic structure or task semantics.  Instead, they rely on surface-level cues in the prompt. 
Here, we analyze EN Matrix--KO Embed setting, where response language varies meaningfully.\footnote{In the Simple Korean-matrix setting, models produce Korean responses in over 99\% of cases, yielding a near-degenerate outcome distribution and rendering cue-based analyses uninformative.}

\paragraph{Script Proportion in the Query.}

\begin{figure}[htb!]
  \centering
  \includegraphics[width=\linewidth]{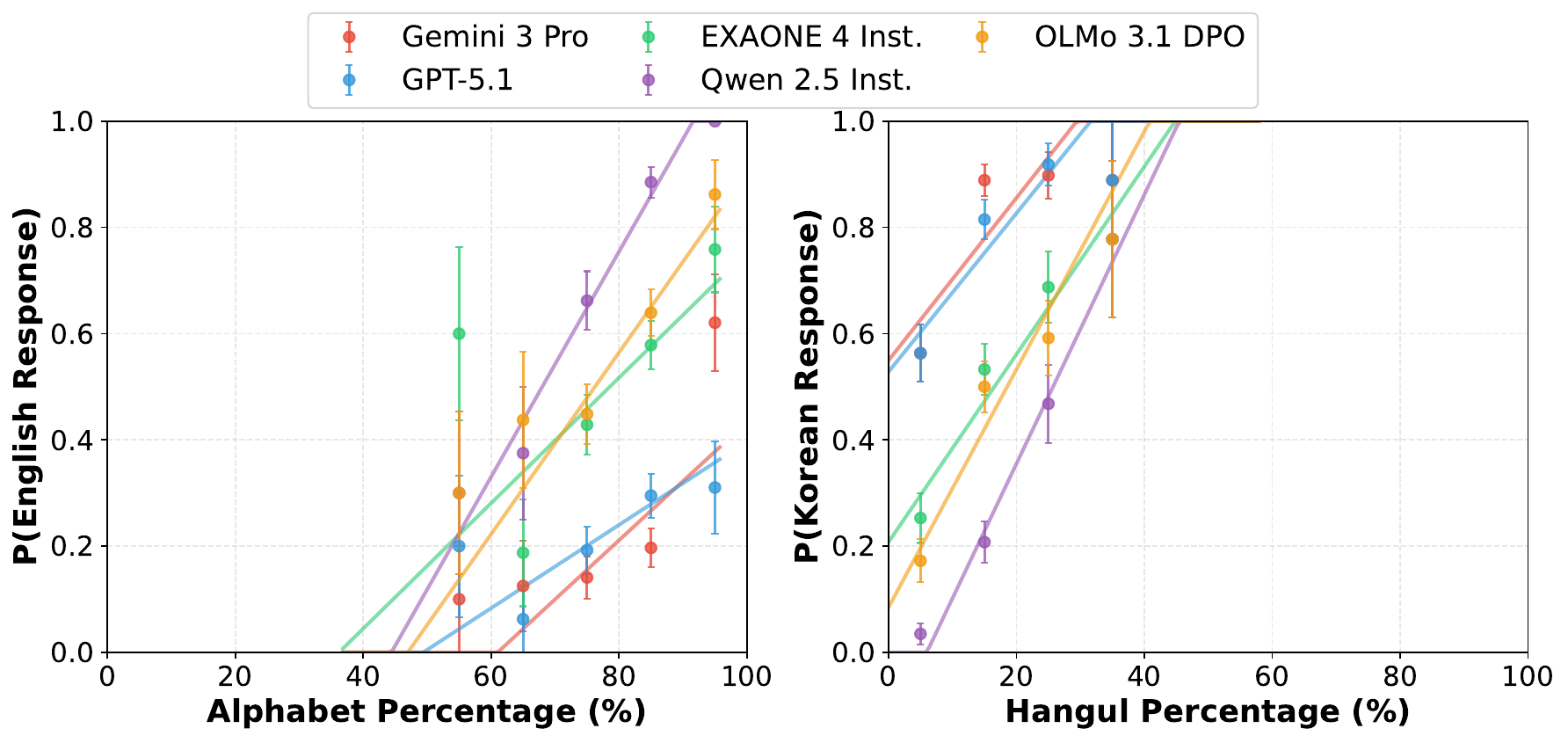}
  \caption{\textbf{Effect of script ratio on output language.} X-axis shows the percentage of alphabetic (left) or Hangul (right) characters in the prompt.}
  \label{fig:script_ratio}
\end{figure}
We find that output language strongly correlates with the script composition of the prompt: English responses increase with the proportion of alphabetic characters, and Korean responses with the proportion of Hangul (Figure~\ref{fig:script_ratio}).
This suggests reliance on coarse script-level statistics rather than deeper linguistic inference.

\paragraph{Language of the Final Word.}
\begin{figure}[htb!]
  \centering
  \includegraphics[width=0.95\linewidth]{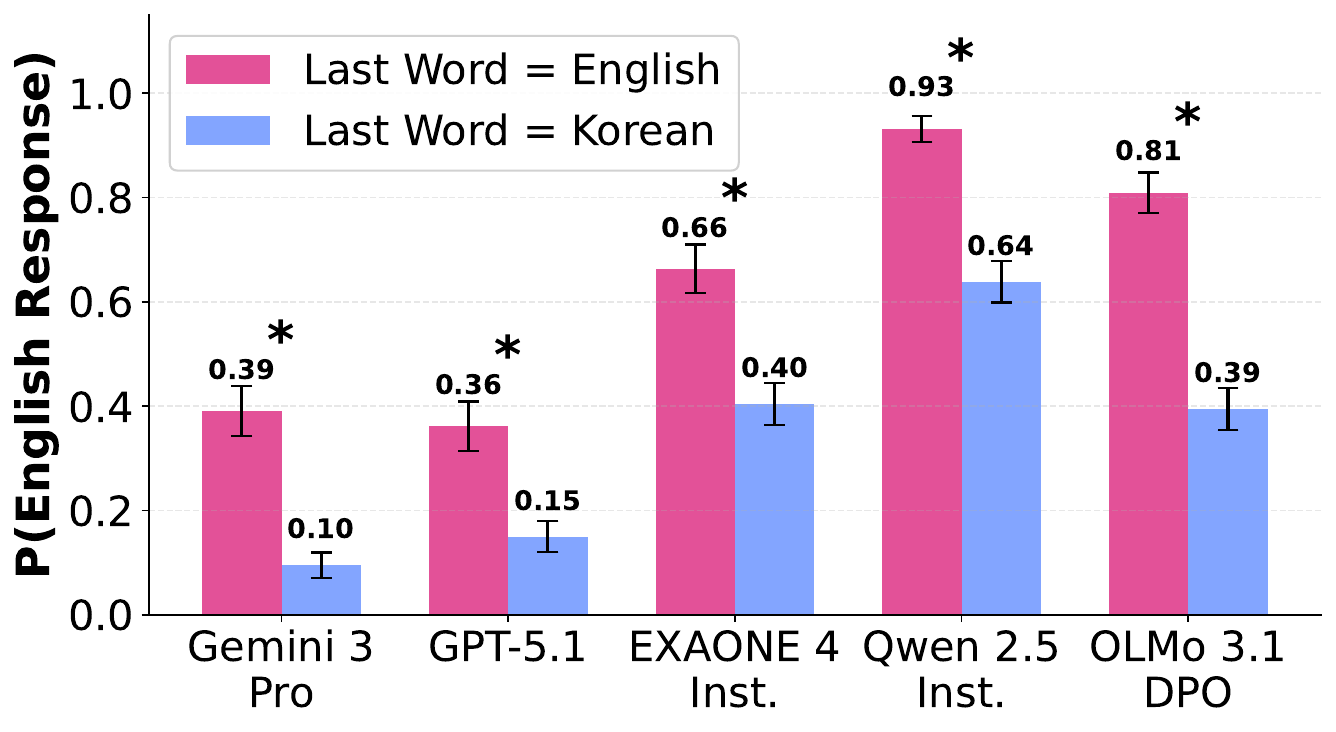}
  \caption{\textbf{Effect of final word language on output language.} Models are significantly more likely to respond in the language of the prompt's final word (all $p$ < 0.001).}
  \label{fig:last_word_effect}
\end{figure}


Beyond global script statistics, localized surface cues also exert strong influence. Figure~\ref{fig:last_word_effect} shows that models are significantly more likely to generate responses in the language of the final word in the prompt, regardless of its syntactic role. By contrast, we find no comparable correlation for the first word (see Appendix~\ref{app:surface_cues}). This effect persists even when the final word appears in a short embedded phrase, indicating a strong recency bias and reinforcing the role of shallow heuristics.

\subsection{Post-training Introduces Language-Specific Biases}
\begin{figure}[htb!]
  \centering
  \includegraphics[width=\linewidth]{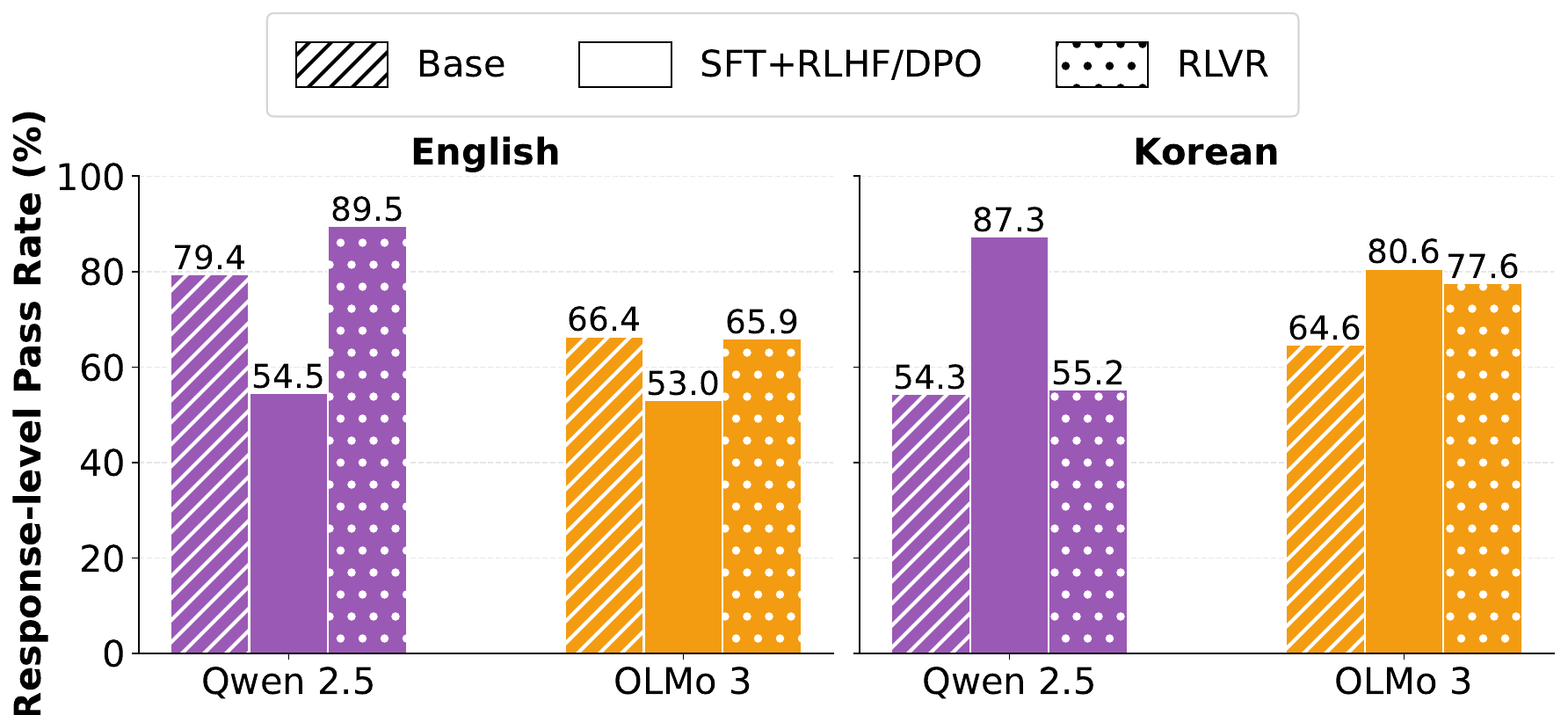}
  \caption{\textbf{Effect of post-training on output language.} Comparison of 32B base models and their post-trained variants across the Qwen-2.5 and OLMo-3 families, aggregated over Simple and Complex settings.}
  \label{fig:posttraining_setting}
\end{figure}

We examine whether post-training improves output language alignment in code-switched contexts by comparing 32B base models with their post-trained counterparts across two model families: \textsc{Qwen 2.5} and \textsc{OLMo-3.1}.
We compare base models, alignment-tuned variants (SFT+RLHF/DPO), and models trained with reinforcement learning from verifiable rewards (RLVR)~\cite{lambert2024tulu} (see Table~\ref{tab:model_training_summary}).

While post-training yields modest overall improvements (see Appendix Figure~\ref{fig:posttraining_language}), the more revealing pattern emerges when disaggregating by expected output language (Figure~\ref{fig:posttraining_setting}). Alignment-tuned models exhibit a pronounced shift toward Korean responses: pass rates drop when English is expected, while Korean response rates increase. This indicates that alignment tuning amplifies a bias toward non-English responses in code-switched contexts.
Interestingly, RLVR models partially reverse this trend, showing higher English response rates than their alignment-tuned counterparts. 
We hypothesize this reflects RLVR's outcome-focused objective, which may favor English as the more reliable output language.

\subsection{Chain-of-Thought Reasoning Does Not Resolve Misalignment}
\label{sec:cot_language}
\begin{figure}[htb!]
  \centering
  \includegraphics[width=\linewidth]{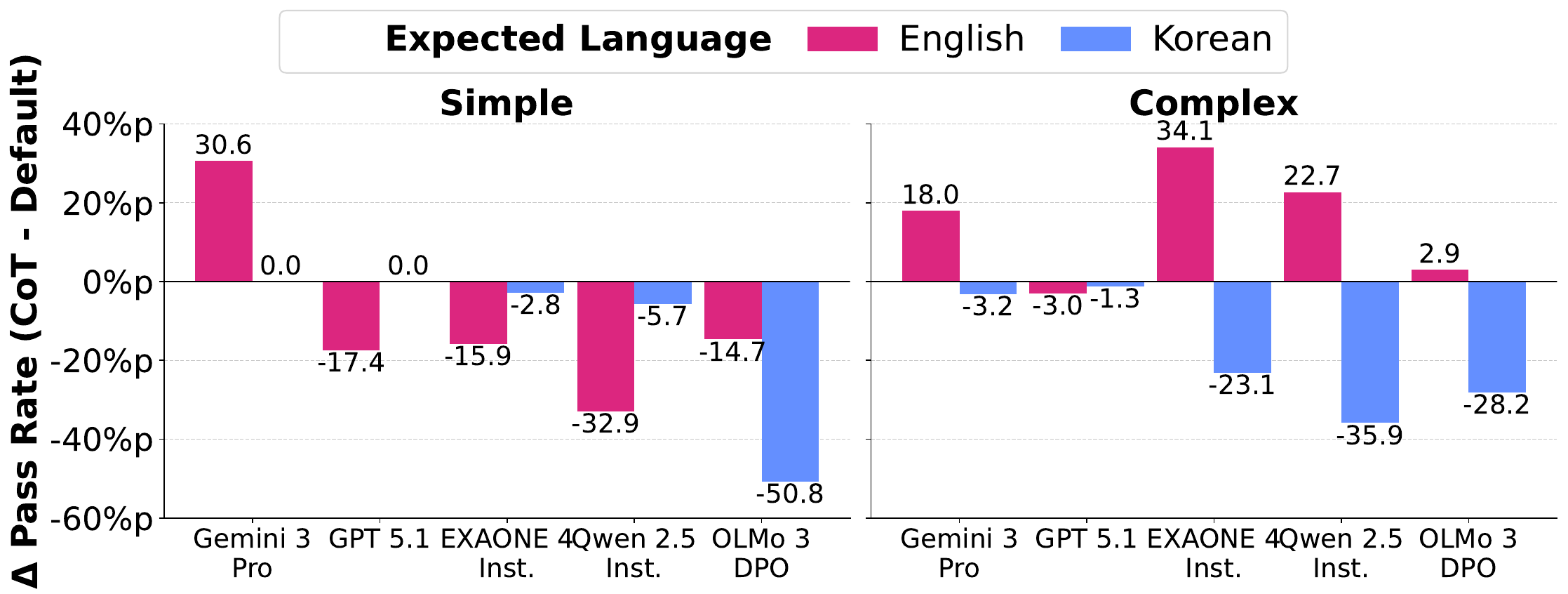}
  \caption{\textbf{Effect of CoT on output language.} Y-axis shows \(\Delta\) Response-level Pass Rate for each model in Simple and Complex settings.} 
  \label{fig:cot_results}
\end{figure}

We apply CoT prompting, instructing models to first decide the output language and then generate the answer. 
Figure~\ref{fig:cot_results} shows the change in response-level pass rate induced by CoT prompting relative to standard prompting. Across both Simple and Complex settings, CoT fails to consistently improve language alignment and often \textit{degrades} performance. Most models exhibit substantial drops in the Simple setting, indicating that explicit reasoning amplifies misalignment rather than mitigating it.

In the Complex setting, CoT-induced deltas become strongly asymmetric: for most models, performance improves for English responses while degrading for Korean responses (and vice versa for \textsc{GPT~5}). This pattern indicates that CoT reshapes how models arbitrate between candidate output languages, but not in a way that reliably aligns with user expectation. 

To understand this failure, we analyze the alignment between the output language selected in the reasoning phase and the language of the final output, using an automated classifier (details in Appendix~\ref{app:cot_method}). 
On average, over 92\% of responses match the language specified in the thought trace (Table~\ref{tab:cot_match_rate}). 
This indicates that performance degradation primarily arises from errors in the \emph{decision} phase rather than the \emph{generation} phase.


Overall, these findings indicate that current LLMs lack reliable internal criteria to find the right output language in code-switched contexts, defaulting to unstable priors despite post-training or reasoning prompts. This leads to our investigation into approaches that explicitly ground or directly align output language preferences.
\label{sec:analysis}

\section{Mitigating Output Language Misalignment}
\normalem
In this section, we explore two strategies to mitigate output language misalignment: (1) Explicit Linguistic Prompting to externally specify language selection criteria at inference time, and (2) Code-Switching Aware DPO to align the model's internal language selection mechanism through targeted training.

\subsection{Explicit Linguistic Instruction Prompting} 
\label{sec:explicit_linguistic_instruction}
Given that models struggle to infer expected output language from implicit cues, we examine whether explicit linguistic guidance at inference time can help. We test zero-shot and few-shot prompting strategies that specify language selection criteria without modifying parameters.

\begin{table}[t]
\centering
\small
\setlength{\tabcolsep}{4pt}
\begin{tabular}{l ccc}
\toprule
Model & Baseline & Zero-shot & 4-shots \\
\midrule
\multicolumn{4}{l}{\textit{\textbf{Simple EN--KO (Expected: English)}}} \\
Gemini 3 Pro & 21.71 & 96.50 & \textbf{98.84} \\
GPT-5.1 & 24.42 & \textbf{41.47} & 16.67 \\
OLMo 3.1 (SFT+DPO) & \textbf{57.36} & 34.11 & 38.37 \\
Qwen 2.5 Inst. & \textbf{75.19} & 25.97 & 2.71 \\
EXAONE 4 Inst.& \textbf{52.34} & 45.74 & 48.45 \\
\midrule
\multicolumn{4}{l}{\textit{\textbf{Complex (KO Inst -- EN Content)}}} \\
Gemini 3 Pro & 69.76 & 66.61 & \textbf{71.26} \\
GPT-5.1 & 71.13 & 71.94 & \textbf{74.70} \\
OLMo 3.1 (SFT+DPO) & \textbf{77.84} & 55.08 & 66.09 \\
Qwen 2.5 Inst & \textbf{60.14} & 56.07 & 55.77 \\
EXAONE 4 & 67.53 & 71.94 & \textbf{77.28} \\
\bottomrule
\end{tabular}
\caption{\textbf{Response-level Pass Rates (\%) under different prompting strategies.} Few-shot prompting can substantially improve performance for some models, but results remain highly model- and setting-dependent.}
\label{tab:prompting_results}
\end{table}

In the zero-shot setting, we prepend a system prompt defining explicit rules: respond in the matrix language for Simple settings, and distinguish instruction vs. content languages for Complex settings. This yields mixed results (Table~\ref{tab:prompting_results}). Large proprietary models benefit substantially---\textsc{Gemini 3} improves from 21.7\% to 96.5\% in Simple (EN Matrix)---while open models show limited or negative effects, with \textsc{Qwen 2.5} dropping from 75.2\% to 26.0\%.

Few-shot prompting augments the system prompt with four demonstrations but again yields inconsistent effects. While some models improve in Complex settings (\textsc{Exaone 4}: 67.5\% → 77.3\%), others degrade in Simple settings (\textsc{Qwen 2.5}: 26.0\% → 2.7\%). This instability suggests that in-context examples cannot reliably override underlying language priors. Notably, \textsc{Gemini 3}  shows consistent gains across both zero-shot and few-shot prompting, aligning with its CoT improvements (Section~\ref{sec:cot_language}).
See Appendix~\ref{app:zero_shot_few_shot} for the prompts used.

Overall, explicit linguistic prompting partially alleviates errors but remains model-dependent and fragile, indicating inference-time prompting alone does not offer a robust solution.



\subsection{Code-Switching Aware DPO (CS-DPO)}
Given that explicit prompting proves insufficient for most open models, we investigate whether the underlying alignment gap can be addressed through direct preference optimization~\citep{rafailov2023direct}.
We develop Code-Switching Aware DPO ({CS-DPO}), a targeted alignment stage designed to penalize incorrect language selection.

\paragraph{Code-Switching Preference Dataset.} We construct a synthetic preference dataset derived from LIMA~\citep{zhou2023lima} dataset. 
Preference pairs $(x, y_w, y_l)$ are created by generating code-switched prompts $x$ with either English or Korean as the matrix language and sampling responses from the baseline model. The \textit{chosen} response $y_w$ matches the prompt's matrix language, while the \textit{rejected} response $y_l$ does not. When a correct response is not naturally produced, $y_w$ is generated using explicit language instructions.
In total, we obtain 1,079 preference pairs for {CS-DPO} training. 
We perform DPO training on top of \textsc{OLMo 3.1 DPO}. Detailed training configurations are provided in Appendix~\ref{sec:appendix_training}.

\begin{table}[t]
\centering
\small
\setlength{\tabcolsep}{4pt} 
\begin{tabular}{l c l}
\toprule
Setting / Metric & Baseline & Ours (+\textsc{CS-DPO}) \\
\midrule
\multicolumn{3}{l}{\textit{\textbf{EN-Ko Simple}}} \\
EN Matrix--KO Embed & 57.36 & \textbf{66.28} \inc{8.92} \\
KO Matrix--EN Embed & 99.29 & \textbf{99.65} \inc{0.36} \\
\midrule
\multicolumn{3}{l}{\textit{\textbf{EN-Ko Complex (Unseen)}}} \\
EN Inst--KO Content    & \textbf{47.90} & 43.77 \dec{-4.13} \\
KO Inst--EN Content    & 77.84 & \textbf{83.96} \inc{6.12} \\
\midrule
\multicolumn{3}{l}{\textit{\textbf{EN-Zh Simple (Unseen)}}} \\
EN Matrix--ZH Embed & 32.47 & \textbf{49.07} \inc{16.60} \\
ZH Matrix--EN Embed & \textbf{97.42} & 95.94 \dec{-1.88} \\
\bottomrule
\end{tabular}
\caption{\textbf{Response-level Pass Rate (\%) with CS-DPO.} Values in parentheses denote the performance gap relative to the baseline (SFT+DPO). CS-DPO exhibits significant gains when English is the expected output language, generalizing to unseen complex and cross-lingual settings.}
\label{tab:dpo_results}
\end{table}



\paragraph{Results.}
Table~\ref{tab:dpo_results} presents the impact of {CS-DPO}. The proposed method improves language consistency across most settings. 
In the Simple setting, {CS-DPO} substantially increases the pass rate for EN Matrix--KO Embed queries on our benchmark (54.27\% $\rightarrow$ 62.11\%) while maintaining strong performance on KO Matrix--EN Embed.
We also observe strong cross-lingual transfer: despite training only on EN--KO, {CS-DPO} yields large gains in the unseen EN Matrix--ZH Embed setting (32.47\% $\rightarrow$ 47.97\%), suggesting a language-agnostic alignment.
This generalization also extends to the Complex setting, improving robustness in KO Instruction--EN Content configurations (77.84\% $\rightarrow$ 83.96\%) despite being trained only on Simple (intra-sentential) examples. 

\section{Conclusion}
\normalem
This study systematically examines output language misalignment in code-switched LLM interactions.
We introduce \textsc{OLA}, an English--Korean benchmark spanning intra-sentential mixing and instruction--content mismatches. We show that frontier models exhibit systematic, asymmetric misalignment, often defaulting to non-English outputs, and that such failures cannot be resolved through inference-time reasoning alone.
We further show that these errors primarily stem from insufficient alignment: a small amount of targeted preference alignment (CS-DPO) substantially improves language-aligned outputs without explicit language instructions. While \textsc{OLA} focuses on English--Korean, our findings highlight a broader need for multilingual LLMs to align with users' implicit expectations in code-switched interactions.


\section*{Limitations}
Our study primarily focuses on English--Korean code-switching, with additional Simple intra-sentential evaluations on English--Chinese and English--Indonesian. Extending the benchmark to more language pairs was constrained by the need for native-speaker expertise to verify the linguistic plausibility of synthesized code-switched text and to validate expected output language preferences. While the authors verified that the expected matrix language was correctly realized in the additional language pairs, full human validation was not conducted due to limited resources. Extending the benchmark across more language pairs remains an important direction for future work.

Second, our analysis is limited to single-turn interactions and does not examine how output language selection evolves over multi-turn dialogue. We also do not conduct a fine-grained interpretive analysis of the mechanisms underlying different types of language errors, such as response-level language selection failures versus unintended word- or character-level language intrusions that violate the expected output language. Finally, we do not explicitly analyze cases where annotators disagree on the expected output language, which reflects the inherent pragmatics of multilingual communication and warrants further study.

\section*{Ethical Considerations}
All human annotation in this work was conducted with appropriate compensation and informed consent. 
Annotators were compensated at 1.5× the local minimum wage, and each annotation task required at most 2.5 hours.
All data used in this work are derived from existing public datasets and are synthetically transformed into code-switched prompts for evaluation. As the transformations preserve semantic content without introducing sensitive attributes, we do not anticipate any direct harm arising from the use of these data.
We use ChatGPT, Gemini, and Cursor for writing and coding assistance.


\normalem
\bibliography{custom}

\clearpage
\section*{Appendix}
\appendix
\section{Representative Queries}
\label{sec:representative_template_queries}

\begin{table}[ht]
\centering
\small
\setlength{\tabcolsep}{4pt}
\begin{tabularx}{\columnwidth}{X c}
\toprule
Query & Expected Lang. \\
\midrule
Explain in simple terms the following content & Instruction \\
Explain this to a beginner, what is the concept, what is it trying to say & Instruction \\
In the passage provided, what is the prediction? & Instruction \\
Please write the following in a legal way & Content \\
Please recompose this with more details & Content \\
Please draft a reply to the update above. & Content \\
\bottomrule
\end{tabularx}
\caption{Representative sample queries from the dataset. Only a subset of queries is shown due to space limit.}
\label{tab:representative_query}
\end{table}

Table~\ref{tab:representative_query} lists examples of English Instruction and their expected output languages. 
The full list of representative English instruction templates is available on GitHub.

\section{Model Failure Patterns}
\label{sec:failure_patterns}
\subsection{Qualitative Analysis of Failure Patterns}
\paragraph{Qualitative Failure Examples.}
Table~\ref{tab:failure_sample} provides representative examples corresponding to the failure types discussed in Section~\ref{sec:failure_pattern}.
For each error category, we show a minimal input--output pair illustrating how language deviations manifest at the character, sentence, or mid-response level in the Simple setting.

\begin{table*}[htb!]
\small
\begin{tabularx}{\linewidth}{l p{0.27\linewidth} X p{0.13\linewidth}}
\toprule
\# & Query & Response & Error Type \\ 
\midrule
1 & What are some of the key advancements in 인공지능 that have significantly impacted 의료 and 금융 in the last decade? \newline
(Translation: \emph{What are some of the key advancements in AI that have significantly impacted healthcare and finance in the last decade?}) &
Here are some of the most influential AI (인공지능) advances over roughly the last 10 years, focusing on 의료 and 금융.  \textcolor{blue}{---  \#\# 1. 공통 기반 기술 (의료·금융 모두에 영향)  \#\#\# 딥러닝(Deep Learning) 혁신 - **CNN, RNN, LSTM, Transformer** 등 모델 구조 발전   - 대규모 GPU/TPU 연산과 클라우드 인프라 확산   → 복잡한 패턴 인식, 예측, 자연어 처리 성능이 비약적으로 향상되어 의료 영상 분석, 챗봇, 리스크 모델링, 이상 징후 탐지 등에 직접 활용. [\ldots]} (\texttt{GPT 5.1}) &
Wrong From Middle \\ 
\midrule
2 & 다양한 문화나 국가에서 \textcolor{blue}{Liberation Day}의 중요성은 무엇인가요? \newline
(Translation: \emph{What is the importance of Liberation Day in various cultures and countries?}) &
\textcolor{blue}{Liberation Day}는 많은 나라에서 역사적으로 중요한 자유를 기념하는 데 사용되는 특정 날을 말합니다. 이 날은 각 나라마다 역사적 맥락과 의미가 다릅니다. 여기 몇 가지 주요 예시를 드릴게요:  1. **한국 – 광복절 (8월 15일)**      한국에서는 1945년 8월 15일에 일본이 대한민국을 \textcolor{blue}{колон}(Pronunciation: \emph{kolon})\textcolor{blue}{리}(Pronunciation: \emph{li})\textcolor{blue}{ализ}(колониализм---Russian for \emph{colonialism}; a tokenization error produces a mixed-script
form, but the intended pronunciation and meaning remain \emph{colonialism}, which makes sense in this context.) 에서 독립시켰을 때를 '광복절'이라고 불리는 자유의 날로 기념합니다.  [\ldots]
 (\texttt{OLMo 3.1 DPO}) &
Char Level Error \\ 
\midrule
3 & Past tense를 사용할 때 이야기의 서술 방식에서 다른 시제와의 주요 differences는 무엇인가요? \newline
(Translation: \emph{What are the main differences in narrative style when using the past tense compared to other tenses?}) &
Past tense(과거 시제)를 사용할 때, 이야기의 서술 방식에서 다른 시제와의 주요 차이점은 다음과 같습니다:  1. **시간적 배경**: 과거 시제는 과거의 사건이나 행동을 표현합니다. 이는 현재 시제(현재의 사건이나 행동을 표현)나 미래 시제(미래의 사건이나 행동을 예상하거나 계획)와는 시간적 배경이 다릅니다.  2. **사용하는 동사 형태**: 영어에서 과거 시제는 주어에 상관없이 일반적으로 동사의 기본 형태에 'ed'를 붙여 표현하거나 불규칙 동사의 과거 시제 형태를 사용합니다.  [\ldots] 3. **시점의 이동**: 과거 시제는 이야기의 흐름이 이미 완료된 사건이나 행동으로 이동함을 나타냅니다. 따라서, 과거 시제는 이야기의 진행이 완료된 상태를 나타내는 데 사용되며, 이는 현재 시제나 미래 시제와는 다른 이야기의 진행 방식을 보여줍니다.  4. **\textcolor{blue}{나ARRy;的叙述方式与其它时态的主要差异在于时间背景、动词形式的使用、以及叙述进程的移动方式。希望这些解释对您有所帮助！如果有更多问题，请随时提问。} 
 (\texttt{Qwen 2.5 Instruct}) &
Sent Level Error \\ 
\midrule
4 & AI autonomous 무기의 주요 ethical 고려 사항에 대한 blog 게시물 작성하기 \newline
(Translation: \emph{Write a blog post discussing the key ethical considerations of AI autonomous weapons.}) &
\# **AI 자율 무기의 윤리적 고려 사항: 우리는 어디까지 허용해야 할까?** \textcolor{blue}{인공지능(AI)} 기술이 급속히 발전하면서 **\textcolor{blue}{자율 무기 시스템(Autonomous Weapons Systems, AWS)}**에 대한 논의도 뜨겁게 이어지고 있습니다. [\ldots] (\texttt{Exaone}) &
Correct / \newline Excessive bilingual notations \\ 
\midrule
5 & Could you give me some in depth tips to make it go viral and spread awareness widely. \newline 저는 사람들이 하루 동안 기술 없이 지내도록 장려하는 바이럴 운동을 만들고 싶습니다. 목표는 마음챙김과 실제 세상과의 연결을 촉진하는 것입니다. 참여를 추적하고 이야기를 공유하기 위해 해시태그를 사용할까 합니다. 이 아이디어를 효과적으로 확산시키고 사람들이 하루 동안 연결을 끊는 것에 대해 흥미를 느끼도록 하려면 어떻게 해야 할까요? & [\ldots] ---  \#\#\# **2. 참여 유도를 위한 창의적인 캠페인 설계** - **\textcolor{blue}{도전 과제} (Translation: \emph{challenge}; excessive word-for-word translation; should be replaced with transliteration (\emph{챌린지})) 형식**으로 제안:     "하루 동안 스마트폰을 내려놓고, 대신 책을 읽거나 산책을 해보세요. 경험을 공유해 주세요! [\ldots] \newline ---  \#\#\# **3. 인플루언서 및 커뮤니티 활용** - **\textcolor{blue}{소셜 미디어} (Translation: \emph{Social media}; should be replaced with \emph{SNS}, an English loanword that have been appropriated in English and not readily understandable to English native speakers) 인플루언서**와 협업:     \textcolor{blue}{마인드풀니스} (Translation: \emph{mindfulness}; should be replaced with \emph{마음챙김} rather than awkward transliteration), 웰빙, 지속 가능한 라이프스타일 분야의 인플루언서가 캠페인을 홍보하도록 제안하세요.   - **지역 커뮤니티/단체**와 연계:     도서관, 카페, 공원 등에서 오프라인 이벤트를 개최해 참여자를 모으세요. [\ldots] (\texttt{Exaone}) &
Correct / \newline Over-translation / Inappropriate lexical choice \\ 
\bottomrule
\end{tabularx}
\caption{Failure samples generated by LLMs}
\label{tab:failure_sample}
\end{table*}




\paragraph{Unexpected Language Choices.}
Beyond the response-level pass rate, our qualitative analysis reveals more subtle forms of language confusion. In many instances, models produce responses that are not monolingual but are themselves code-switched, often in ways that are unnatural or erroneous. We identify three common patterns as follows. Table~\ref{tab:failure_sample} rows 2--5 provide illustrative examples of these failure types, demonstrating that even when a model's output is not a complete failure in terms of language choice, its ability to maintain linguistic consistency remains a significant challenge.

\begin{enumerate}[leftmargin=*,topsep=0pt,itemsep=2pt,parsep=0pt]
    \item \textbf{Mid-Response Language Switching}: The model begins generating a response in the correct language but abruptly switches to the other language mid-sentence or mid-paragraph without a clear rhetorical reason.
    Interestingly, this language alternation occurs within languages not used in code-switching queries; random languages (\eg Chinese or Cyrillic script) are inserted within a response upon a Korean-English code-switching query, maintaining semantic consistency.
    \citet{zhao2024llama, yoo-etal-2025-code-switching} reported that this phenomenon more frequently occurs in continually pre-trained LLMs for language transfer, and we observe that Qwen 2.5, one of the most multilingual models, specifically includes more mid-response language switching instances than other models.
    \item \textbf{Intra-word Switching}: LLM responses occasionally include intra-word switching based on subword-based tokens (\eg byte-pairs), maintaining its pronunciation continuously.
    The inserted languages are random as mid-response language switching.
    This phenomenon only occurs when the model responds in Korean.
    It implies that LLM tokenizers may internally process cross-lingual alignment based on phonemic representation~\cite{jung-etal-2024-mitigating}.
    \item \textbf{Excessive use of code-switching phrases or bilingual notations}: LLMs tend to excessively use bilingual notations in Korean-English or Korean-Hanja (\ie Chinese script used to write Korean) upon code-switching queries. In addition, they tend to repeat phrases from queries in embedded languages in their responses.
\end{enumerate}

\paragraph{English-style Korean.} In Korean responses, both inter- and intra-sentential code-switching queries elicit more use of awkward transliteration words from English rather than Korean phrases (\eg 마인드풀니스) or loanwords that have been appropriated into Korean (\eg 소셜 미디어) (Table~\ref{tab:failure_sample}, row 5).
On the other hand, LLMs also use awkward, excessive word-for-word translations rather than naturally-sounding transliterations (\eg 도전 과제).
In general, LLMs tend to respond in Korean to code-switching queries with translationese, simply converting their internal English generations into word-for-word translations~\citep{zhao2024how, etxaniz-etal-2024-multilingual, zhong-etal-2025-language, bafna2025translation, schut2025multilingual}.

\subsection{Character-Level Language Intrusions Across Alignment Variants}
\begin{table*}[t]
\small
\centering
\begin{tabular}{lcccl}
\toprule
Backend & Correct (\%) & Incorrect (\%) & \#Errors & Most Common Offending Languages \\ 
\midrule
\textsc{Olmo-Base} & 99.76 & 0.24 & 2  & JA (2) \\
\textsc{Olmo-SFT}  & 98.12 & 1.88 & 16 & RU (10), JA (4), HI (2), AR (1) \\
\textsc{Olmo-SFT+DPO}  & 92.23 & 7.77 & 66 & RU (36), HI (13), JA (12), AR (8), TH (1) \\
\textsc{Olmo-Instruct}      & 89.75 & 10.25 & 87 & RU (55), JA (23), HI (13), AR (8) \\
\midrule
\textsc{Qwen-Base} & 99.76 & 0.24 & 2  & RU (1), JA (1) \\
\textsc{Qwen-Instruct}      & 98.00 & 2.00 & 17 & RU (10), JA (5), TH (2), AR (2) \\
\textsc{QWQ}       & 90.58 & 9.42 & 80 & RU (41), JA (26), TH (10), AR (5), HE (3), HI (1) \\
\bottomrule
\end{tabular}
\caption{Character-level language intrusions across model families in the Simple (Korean matrix) setting (total: 283). We report the number and distribution of incorrect cases involving unexpected scripts. Chinese characters are excluded due to potential confounds with Sino-Korean orthography.}
\label{tab:char_level_intrusions}
\end{table*}

\begin{table*}[t]
\small
\centering
\begin{tabular}{lcccl}
\toprule
Backend & Correct (\%) & Incorrect (\%) & \#Errors & Most Common Offending Languages \\ 
\midrule
\textsc{Olmo-Base} & 99.62 & 0.38 & 3  & TH (1), JA (1), RU (1) \\
\textsc{Olmo-SFT}  & 99.36 & 0.64 & 5  & RU (4), HI (1) \\
\textsc{Olmo-SFT+DPO}  & 96.55 & 3.45 & 27 & RU (16), JA (7), AR (3), HI (2), TH (1), HE (1) \\
\textsc{Olmo-Instruct} & 96.55 & 3.45 & 27 & RU (13), JA (9), AR (4), HI (3) \\
\midrule
\textsc{Qwen-Base} & 99.74 & 0.26 & 2  & HI (1), RU (1) \\
\textsc{Qwen-Instruct}      & 99.74 & 0.26 & 2  & TH (1), JA (1) \\
\textsc{QWQ}       & 99.49 & 0.51 & 4  & RU (4) \\
\bottomrule
\end{tabular}
\caption{Character-level language intrusions across model families in the Simple (English Matrix) setting (total: 258). We report the number and distribution of incorrect cases involving unexpected scripts. Chinese characters are excluded due to potential confounds with Sino-Korean orthography.}
\label{tab:char_level_intrusions_en_matrix}
\end{table*}

Tables~\ref{tab:char_level_intrusions} and~\ref{tab:char_level_intrusions_en_matrix} summarize character-level language intrusions in the Simple setting for Korean-matrix and English-matrix prompts, respectively.
We report the proportion and distribution of incorrect cases in which models produce characters from unexpected scripts.
Across both matrix languages, base models exhibit near-perfect behavior, while post-trained variants show a substantially higher incidence of such intrusions.
In particular, alignment-tuned and instruction-following models introduce characters from unrelated languages (\eg Russian, Japanese, Hindi, Arabic), suggesting increased susceptibility to fine-grained language leakage after post-training.
Chinese characters are excluded from this analysis due to potential confounds with Sino-Korean orthography.

\section{Models and Experimental Setup}
\subsection{Models}
\label{app:main_experimental_setting}
We evaluate a diverse set of multilingual large language models (LLMs), including both proprietary and open-weight models:
\begin{itemize}[itemsep=0.2em, topsep=0.2em]
    \item \textbf{Gemini 3}: Gemini 3 Pro~\cite{google2025gemini3promodelcard}
    \item \textbf{GPT-5.1}: GPT-5.1~\cite{openai2025gpt5systemcard}~\footnote{version: gpt-5.1-2025-11-13}
    \item \textbf{Qwen 2.5}: Qwen 2.5 Instruct 32B~\cite{qwen2025qwen25}~\footnote{\url{https://huggingface.co/Qwen/Qwen2.5-32B-Instruct}}
    \item \textbf{Exaone 4}: Exaone 4.0.1 32B~\cite{research2025exaone4}~\footnote{\url{https://huggingface.co/LGAI-EXAONE/EXAONE-4.0.1-32B}}.
    \item \textbf{OLMo 3.1 DPO}: OLMo 3.1  32B Instruct DPO~\cite{olmo2025olmo}~\footnote{\url{https://huggingface.co/allenai/Olmo-3.1-32B-Instruct-DPO}}.
    \item \textbf{Qwen 2.5 Base}: Qwen 2.5 Base~\cite{qwen2025qwen25}~\footnote{\url{https://huggingface.co/Qwen/Qwen2.5-32B}}.
    \item \textbf{QwQ}: QwQ 32B~\cite{qwen2025qwq32b}~\footnote{\url{https://huggingface.co/Qwen/QwQ-32B}}.
    \item \textbf{OLMo 3 Base}: OLMo 3 32B Base~\cite{olmo2025olmo}~\footnote{\url{https://huggingface.co/allenai/Olmo-3-1125-32B}}.
    \item \textbf{OLMo 3.1 Instruct}: Olmo 3.1 32B Instruct~\cite{olmo2025olmo}~\footnote{\url{https://huggingface.co/allenai/Olmo-3.1-32B-Instruct}}.
\end{itemize}

All open-weight models are 32B-parameter variants.  
For model families with multiple post-training variants (\eg OLMo and Qwen), we additionally evaluate their corresponding base and alternative alignment versions.  
A summary of the evaluated model variants and their post-training or alignment strategies is provided in Table~\ref{tab:model_training_summary}.

\begin{table}[htb!]
\centering
\small
\begin{tabularx}{\linewidth}{p{0.14\linewidth} l X}
\toprule
Model Family & Model Variant & Post-training Method \\ 
\midrule
Qwen & Qwen~2.5~Base & Pretrained (Base) \\
Qwen & Qwen~2.5~Instruct & SFT + RLHF \\
Qwen & QwQ & RLVR \\ 
\midrule
OLMo & OLMo~3~Base & Pretrained (Base) \\
OLMo & OLMo~3.1~Instruct-DPO & SFT + DPO \\
OLMo & OLMo~3.1~Instruct & SFT + DPO + RLVR \\
\bottomrule
\end{tabularx}
\caption{Models and post-training strategies evaluated in this work. All models are 32B parameter variants.}
\label{tab:model_training_summary}
\end{table}

\subsection{Inference Setting}
We set the parameters for all models to: temperature = 0.7, top\_p = 0.9.
4 Quadro RTX 8000 48GB, 2 NVIDIA H200 141GB were used with CUDA version 12.4 when running open-sourced Models EXAONE, Qwen 2.5 Instruct 32B, and OLMo 3.1 DPO 32B.

\section{Additional Analyses}
\subsection{Surface Cues}
\label{app:surface_cues}
\paragraph{Relationship between the language of the first word in the prompt and response language.}
\begin{figure}[htb!]
  \centering
  \includegraphics[width=\linewidth]{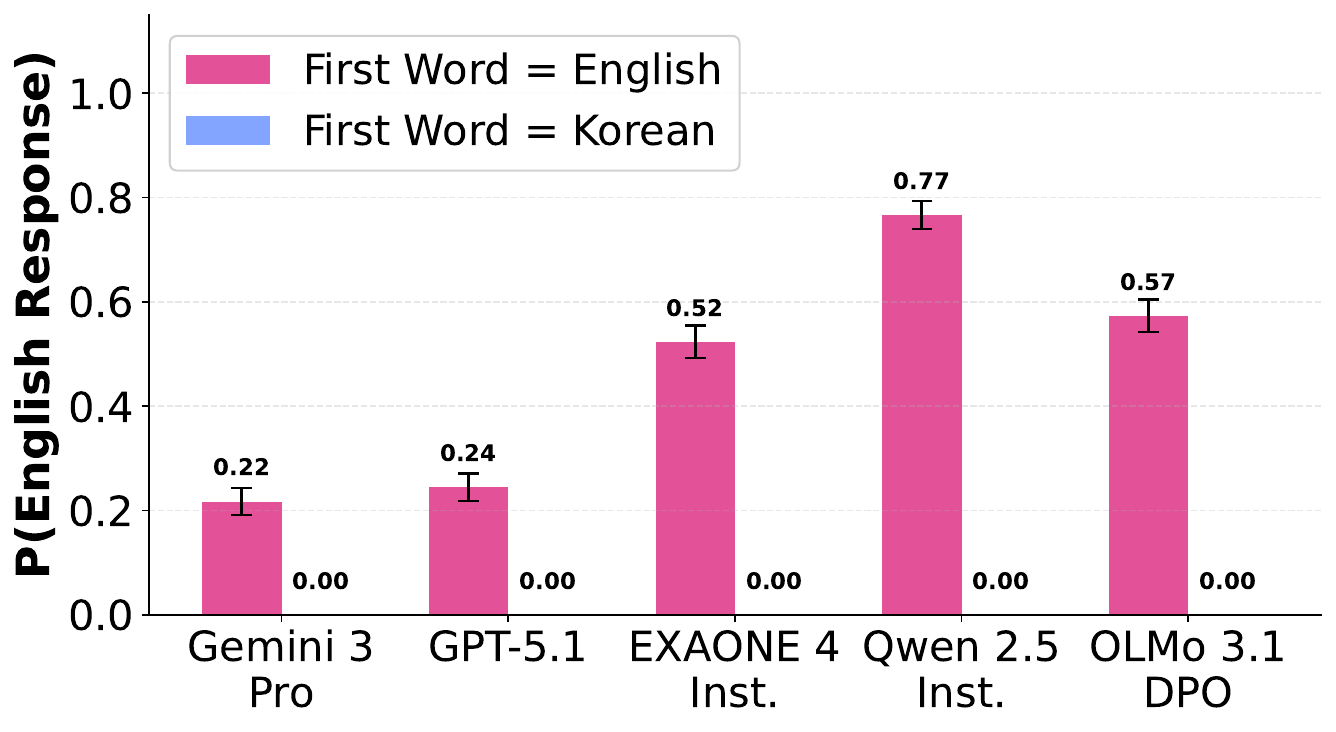}
  \caption{Effect of First Word Language on Output Language.}
  \label{fig:first_word_effect}
\end{figure}
In contrast to the last word effect (Figure~\ref{fig:last_word_effect}, models' response language does not correlate with the first word of the query.

\paragraph{Instruction Position in the Complex Setting.}


\begin{figure}[htb!]
  \centering

  \begin{subfigure}[t]{\linewidth}
    \centering
    \includegraphics[width=\linewidth]{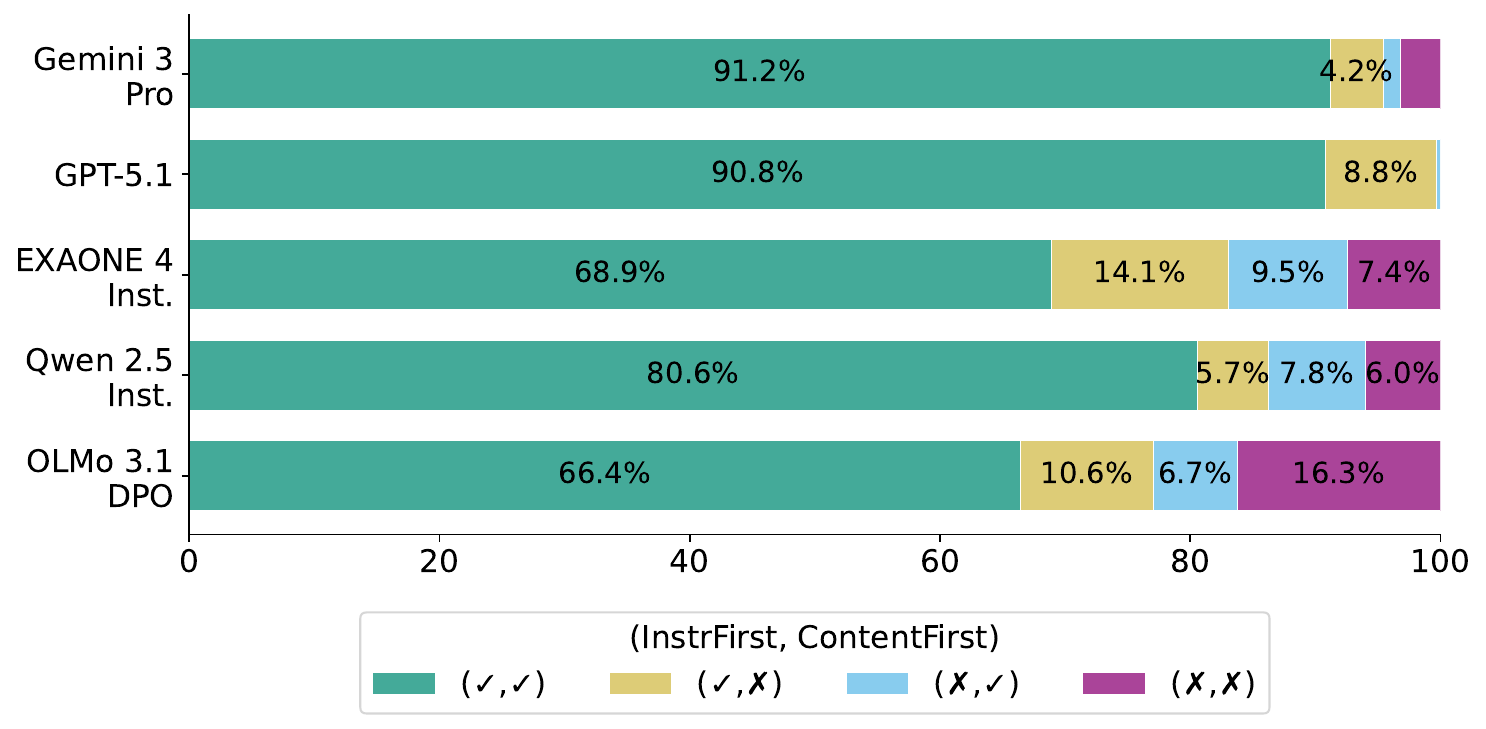}
    \caption{Expected output language: Korean}
    \label{fig:inst_content_order_ko}
  \end{subfigure}

  \vspace{0.6em}

  \begin{subfigure}[t]{\linewidth}
    \centering
    \includegraphics[width=\linewidth]{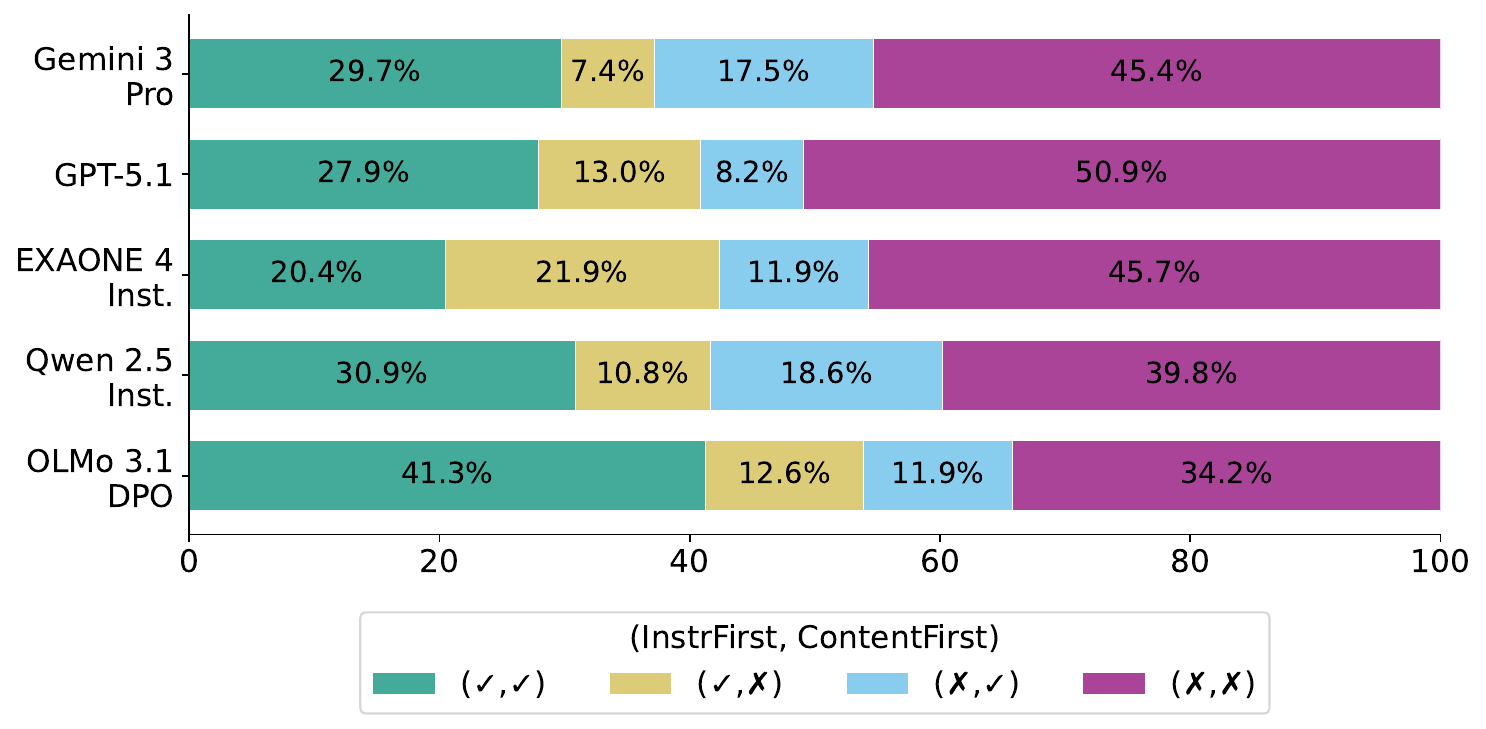}
    \caption{Expected output language: English}
    \label{fig:inst_content_order_en}
  \end{subfigure}

  \caption{
  Effect of instruction position (InstrFirst vs.\ ContentFirst) on model Response-level Pass Rate in complex code-switched prompts, stratified by expected output language.
  Each bar shows the proportion of test cases where models answered correctly under both orderings (\cmark,\cmark), only when instruction came first (\cmark,\xmark), only when content came first (\xmark,\cmark), or neither (\xmark,\xmark).
  }
  \label{fig:inst_content_order}
\end{figure}

Figure~\ref{fig:inst_content_order} analyzes the effect of instruction position (InstrFirst vs. ContentFirst) under complex code-switched prompts, conditioning on the expected response language.
When the expected language is Korean, all models exhibit high robustness to instruction ordering, with the majority of examples correctly handled regardless of whether the instruction appears before or after the content.
When English is expected, the overall pass rate drops substantially across models, reflected by a larger fraction of cases where neither ordering yields a correct response. However, among instances where at least one ordering succeeds, the relative robustness to instruction position remains broadly comparable to the Korean-expected setting.




\subsection{Chain-of-Thought Language Alignment Analysis}
\label{app:cot_method}

\begin{table}[tp]
\centering
\small
\begin{tabular}{l cc}
\toprule
Model & \begin{tabular}[c]{@{}c@{}}EN Matrix\\--KO Embed\end{tabular} & \begin{tabular}[c]{@{}c@{}}KO Matrix\\--EN Embed\end{tabular} \\
\midrule
GPT 5.1 & 93.17 & 100.00 \\
Gemini 3 & 98.63 & 99.66 \\
EXAONE 4 Inst. & 97.60 & 93.86 \\
Qwen 2.5 Inst. & 95.90 & 90.44 \\
OLMo 3.1 DPO & 96.59 & 58.70 \\
\bottomrule
\end{tabular}
\caption{CoT Language Decision--Response Consistency (\%). This metric measures how often the language of the final response matches the language selected during the intermediate reasoning step.}
\label{tab:cot_match_rate}
\end{table}
We analyze alignment between CoT reasoning and final response language using an LLM-based classifier (Qwen-2.5-Instruct) to extract expected language from thought traces (prompts in Appendix~\ref{app:classification_prompt}). Manual validation of 100 random samples showed zero classification errors.

Table~\ref{tab:cot_match_rate} shows high decision-execution alignment (>90\% for most models), indicating performance degradation stems from incorrect language decisions rather than execution failures.
However, the fact that this alignment remains well below 100\% suggests that even for a seemingly simple decision such as response language choice, residual misalignment between reasoning and execution~\cite{paul-etal-2024-making, chan2025predictalignmentmodelsfinish} persists.

\section{Additional Experimental Results}
\subsection{Multilingual Results}
\label{app:multilingual}

\begin{table*}[t]
\centering
\small
\begin{tabular}{llcc}
\toprule
\textbf{Language Pair} & \textbf{Model} &
\textbf{EN Target (\%)} &
\textbf{Non-EN Target (\%)} \\
\midrule
\multirow{5}{*}{Korean--English}
& Gemini 3 Pro        & 20.15 & \textbf{100.00} \\
& GPT-5.1             & 21.50 & \textbf{100.00} \\
& Qwen 2.5 Instruct   & 71.67 & \textbf{97.27}  \\
& EXAONE 4 Instruct   & 48.99 & \textbf{100.00} \\
& OLMo 3.1 DPO    & 54.27 & \textbf{98.98}  \\
\midrule
\multirow{5}{*}{Chinese--English}
& Gemini 3 Pro        & 36.53 & \textbf{99.25} \\
& GPT-5.1             &  8.80 & \textbf{99.25} \\
& Qwen 2.5 Instruct   & 79.40 & \textbf{98.51} \\
& EXAONE 4 Instruct   & 67.67 & \textbf{74.25} \\
& OLMo 3.1 DPO    & 32.47 & \textbf{97.42} \\
\midrule
\multirow{5}{*}{Indonesian--English}
& Gemini 3 Pro        & 41.57 & \textbf{99.49} \\
& GPT-5.1             & 16.85 & \textbf{97.42} \\
& Qwen 2.5 Instruct   & 44.38 & \textbf{97.94} \\
& EXAONE 4 Instruct   & 67.98 & \textbf{88.66} \\
& OLMo 3.1 DPO    & 31.46 & \textbf{93.81} \\
\bottomrule
\end{tabular}
\caption{Response-level pass rates (\%) in the Simple code-switching setting.
Across language pairs, models show a consistent asymmetric bias toward non-English responses:
performance drops sharply when English is the expected output, while non-English targets are handled reliably.}
\label{tab:multilingual_failure}
\end{table*}

\paragraph{Dataset Construction for Additional Languages.}
To evaluate whether the observed output language misalignment generalizes beyond Korean--English, we extend the Simple setting to Chinese and Indonesian.
For both language pairs, the underlying monolingual English queries are drawn exclusively from the Language Confusion Benchmark~\citep{marchisio-etal-2024-understanding}, without incorporating additional queries from WildChat.
We follow the same data generation procedure as in the Korean--English Simple setting, replacing embedded noun phrases with their Chinese or Indonesian counterparts while preserving the English matrix language.
The authors manually verified that the synthesized prompts correctly realize the intended matrix--embedded language structure; however, due to limited resources, we did not conduct large-scale human validation of expected output language preferences for these additional language pairs.

\paragraph{Raw Pass Rate by Language Target.}
Table~\ref{tab:multilingual_failure} provides the raw response-level pass rates for English-target and non-English-target conditions in the Simple code-switching setting.
These results support the analysis in \autoref{sec:main_overall} by making the underlying asymmetry explicit: performance degradation is driven primarily by failures when English is the expected response, rather than by instability in non-English generation.

\subsection{Post-training}
Figure~\ref{fig:posttraining_setting} presents response-level pass rates disaggregated by the expected output language (English vs.\ Korean), aggregated over both Simple and Complex settings.
Across both the Qwen-2.5 and OLMo-3 families, alignment-tuned variants (SFT+RLHF/DPO) and RLVR-trained models show a consistent shift toward higher pass rates in the Complex setting, while no clear trend appears in the Simple setting. 

\begin{figure}[htb!]
  \centering
  \includegraphics[width=\linewidth]{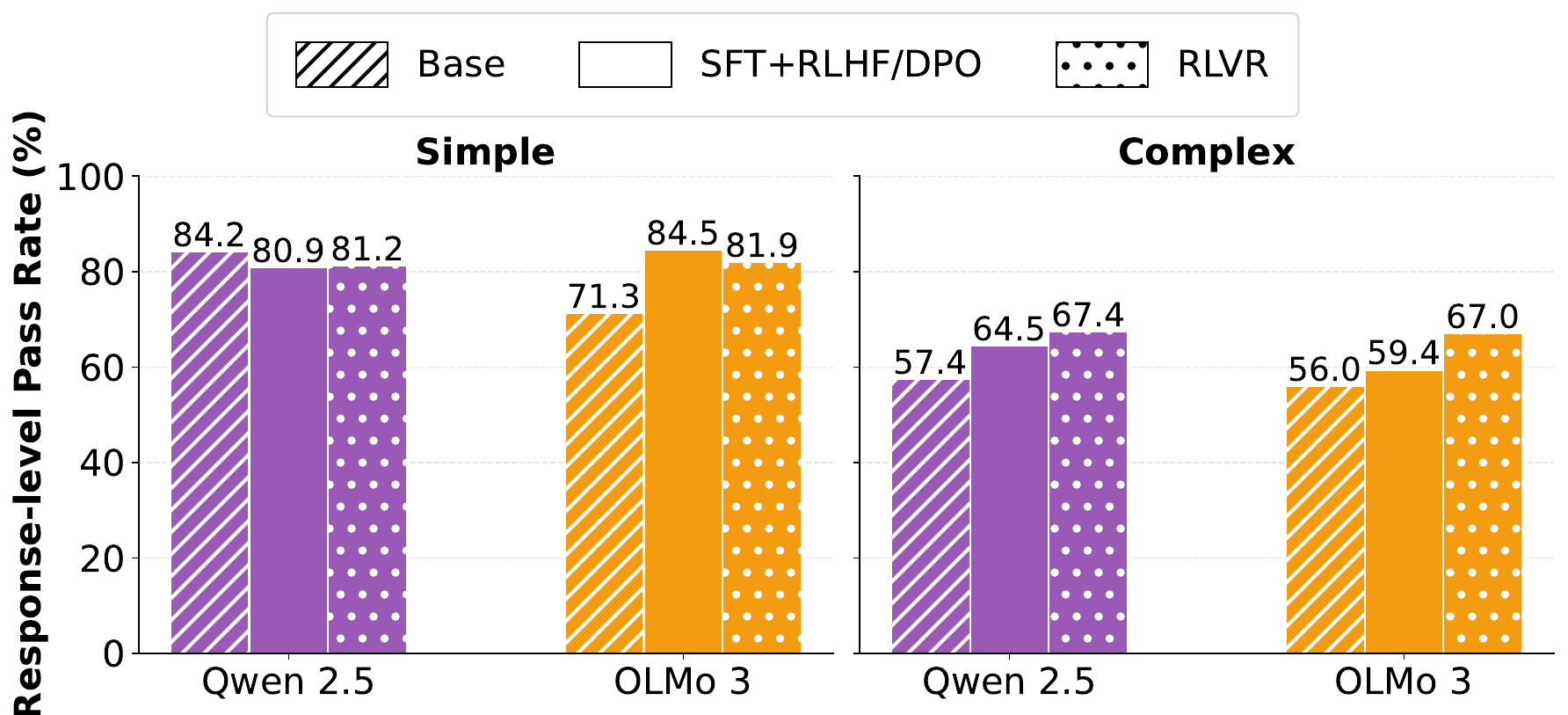}
  \caption{Effect of Post-training in Simple and Complex settings.}
  \label{fig:posttraining_language}
\end{figure}

\section{Post-training Details}
\label{sec:appendix_training}

We conduct additional post-training using Direct Preference Optimization (DPO) to assess whether code-switching failures stem from insufficient alignment data rather than model capacity. All experiments are performed using full-parameter updates on \textsc{OLMo-3.1-32B-Instruct-DPO}.

\paragraph{Training Data.}
The DPO training dataset consists of preference pairs constructed from code-switched queries in the Simple setting. Each pair contains a \emph{chosen} response that follows the intended matrix language and a corresponding \emph{rejected} response that violates it. Both the chosen and rejected responses are derived from the base model, obtained by sampling multiple responses given the code-switched query. This is important to maximize the impact of the DPO training, where both the chosen and rejected responses are actually possible to be generated by the model. Also, the prompt provided to the model contains only the original code-switched query, without any explicit instruction specifying the response language. The dataset is split into training and validation sets.

\paragraph{Training Configuration.}
We perform full-parameter DPO training using the TRL framework. Training is conducted on 8 NVIDIA H200 GPUs 141GB with mixed-precision (\texttt{bf16}). The per-device batch size is set to 1, with gradient accumulation over 8 steps, resulting in an effective batch size of 64. We train for a single epoch using the AdamW optimizer with a cosine learning rate schedule. The learning rate is set to $5 \times 10^{-6}$ with a warmup ratio of 0.03. The DPO temperature parameter $\beta$ is fixed to 0.1.

\paragraph{Sequence Formatting.}
Each training example is formatted into \texttt{prompt}, \texttt{chosen}, and \texttt{rejected} strings using the model’s chat template. To avoid exceeding the model’s context window, we truncate sequences at the token level, allocating up to half of the context length to the prompt and the remainder to the response. 

\paragraph{Optimization and Evaluation.}
Model checkpoints are saved every 200 steps, and evaluation is performed at the same interval on the validation split. To reduce memory overhead, we adopt reference-free DPO and do not maintain a separate frozen reference model. All runs are seeded with a fixed random seed (42) to ensure reproducibility.


\section{Human Annotation Details}
\label{app:human_annotation}

\paragraph{Annotator Recruitment.}
Annotators were recruited via a publicly accessible bulletin board within the authors' institution, which is visible to both students and staff. Participation was voluntary, and all annotators were compensated at a rate of \$10 USD per hour. No personally identifiable information was collected beyond language background relevant to the task.
The annotation was conducted by six native speakers (three English, three Korean). All annotators reported being able to understand and use the embedded language at a basic level, reflecting realistic code-switching users rather than balanced bilinguals.

Annotators were provided with written guidelines describing the task and rating criteria. For each query, annotators were asked to:
(i) identify the expected output language, and
(ii) rate the severity of a language mismatch as \texttt{Trivial}, \texttt{Uncomfortable}, or \texttt{Critical}.
The full annotation guidelines, including definitions and illustrative examples, are provided in the next section.

To ensure reliability, we retain only samples for which at least two annotators agree on the expected output language. Disagreements were not resolved through discussion, as our goal was to capture user-level sensitivity rather than enforce a single ``correct'' judgment.

\subsection{Annotation Guidelines}
\label{app:annotation_guidelines}

Annotators were asked to read user prompts that mix English and Korean and to judge the expected output language and the usability impact of language mismatches.

\paragraph{Task Overview.}
For each prompt, annotators completed two judgments:
(i) selecting the language they would expect the model to respond in, and
(ii) rating how inconvenient it would be if the model responded in a different language.
Annotators were instructed to assume that they themselves authored the prompt and to base their judgments on natural interaction expectations rather than prescriptive rules.

\paragraph{User Prompt.}
Each item consists of a user query intended as input to a language model such as ChatGPT.
Annotators were instructed to read the prompt and judge which language it would be most natural for the model’s response to be in.
Even if the prompt content was not fully understood, annotators were asked not to skip the item as long as the expected response language could be inferred.
For long prompts, annotators were allowed to skim the content and focus only on cues relevant to output language expectation.

\paragraph{Expected Output Language.}
Annotators selected the primary language they would expect the model’s response to be written in.
Other languages may appear in the response, but the selection should reflect the dominant response language. Provided options were: (1) English, (2) Korean, (3) Either (only if the prompt does not clearly favor one language)

\paragraph{Criticality of Language Mismatch.}
Annotators rated how serious it would feel if the model responded in a language different from the expected one, from a usability perspective.

Options were as follows:
\begin{itemize}[leftmargin=*, topsep=2pt, itemsep=1pt]
    \item \textbf{Critical}: The response would be rejected and re-generated with an explicit language instruction.
    \item \textbf{Uncomfortable}: The response would be accepted, but the annotator would specify the language in subsequent interactions.
    \item \textbf{Trivial}: The mismatch would not cause inconvenience and would not prompt a retry.
\end{itemize}

\paragraph{Task Settings.}
Annotators labeled prompts under two distinct settings:

\textbf{Setting 1: Simple (Intra-sentential Code-switching).}
Prompts consist of a single instruction or question containing mixed-language elements.
The model response corresponds directly to answering the prompt.
For example:
\begin{quote}
\small
\texttt{What emotions and 도전 might a man experience when he finds himself in an 낯선 도시?}
\end{quote}

\textbf{Setting 2: Complex (Instruction--Content Code-switching).}
Prompts consist of an instruction and associated content required to carry out the instruction.
The model response includes both helper text (if any) and the content resulting from executing the instruction.
Annotators were instructed to select the expected language based \emph{only} on the language of the resulting content, not the instruction.
For example:
\begin{quote}
\small
\texttt{Can you continue writing a news article based on this information?}

\texttt{\{Content\}}
\end{quote}

\section{Prompts}
\label{sec:system_prompt}

\subsection{Prompts used for OLA Dataset Construction}
\label{app:gen_OLA_prompts}
The following prompts were used with GPT-4o to generate code-switched queries and to create variations of Contents in the Complex setting.

\begin{tcolorbox}[breakable, enhanced, top=1pt, left=1pt, right=1pt, bottom=1pt, title=Prompt for generating code-switched queries]
\small
You are a bilingual rewriting assistant.\\

TASK\\
  • Input  : an English sentence (E) and its Korean translation (K)\\
  • Output : the code-switched version of E\\
            - Replace about {level} percent of NOUNS / NOUN PHRASES in E with their Korean equivalents taken from K\\
            - Keep the original English word order (S-V-O)\\
            - DO NOT add explanations, examples, tags, or extra sentences\\
            - If there is no suitable Korean equivalent, keep the English word\\\\

[EXAMPLE]\\
Input\\
<English>Topic: Using AI to Augment Human Capabilities\\
Explain a common misconception about your topic.\\\\

<Korean>주제: AI를 사용하여 인간의 능력을 증강하기\\
당신의 주제에 대한 일반적인 오해를 설명하세요.\\

Desired Output\\
<Code-Switch>\\
주제: Using AI to 증강 Human Capabilities\\
Explain 일반적인 오해 about your 주제.\\\\

[BEGIN TASK]\\ 
<English>{question}\\
<Korean>{translation}\\
\end{tcolorbox}

\begin{tcolorbox}[breakable, enhanced, top=1pt, left=1pt, right=1pt, bottom=1pt, title=Prompt for generating variations of existing content]
\small
You are an expert data augmentation assistant. \\\\
    You will be given an existing Instruction and its current Content that together form a user query.\\
    Your task is to invent FOUR NEW Content paragraphs that satisfy ALL of the following conditions:\\
    1. When combined with the SAME Instruction they should form a sensible, coherent query.\\
    2. Each new Content must be DIFFERENT from the original Content and from each other. Do not simply paraphrase, instead be creative. You should use different topics and styles.\\
    3. Each new Content must be BETWEEN 200 and 600 characters (inclusive).\\
    4. Do NOT answer the Instruction — you are ONLY creating new Content, not responses.\\
    5. Do NOT mention these guidelines or any numbering in the output.\\
    Return ONLY a JSON array of the four new Content strings.\\
\textnormal{[CONTEXT]}\\
    Instruction: \{instruction\}\\
    Original Content:\{original\_content\}\\
\textnormal{[END CONTEXT]}\\\\
    \#\#\# OUTPUT FORMAT\\\textnormal{[ "content1 ...", "content2 ...", "content3 ...", "content4 ..." ]}\\
\end{tcolorbox}

\subsection{Chain-of-Thought (CoT) Instruction}
\label{app:cot_prompt}

We used the following prompt to elicit CoT reasoning in the English-matrix setting (Section~\ref{sec:cot_language}). The prompts for the Korean setting follow the same structure, translated into Korean to match the matrix language of the query.

\begin{tcolorbox}[breakable, enhanced, top=1pt, left=1pt, right=1pt, bottom=1pt, title=Chain-of-Thought Prompt (English Matrix)]
\small
First think about which language you should respond in, and then generate the answer.\\\\
Output your response in the following JSON structure:\\\\
\{\\
\hspace*{1em} "thought": "<The language you decided to answer in, and a short explanation>",\\
\hspace*{1em} "answer": "<The final answer to the query>"\\
\}
\end{tcolorbox}

\subsection{Language Decision Classification Prompt}
\label{app:classification_prompt}

The following prompt was used with GPT-4o to classify and extract the expected response language from the generated thought traces of LLMs in CoT-settings (Section~\ref{sec:cot_language}).

\begin{tcolorbox}[breakable, enhanced, top=1pt, left=1pt, right=1pt, bottom=1pt, title=Prompt for Language Decision Classification]
\small
You are a language classifier. Analyze the following text where a model explains which language it decided to respond in.\\\\
Text to analyze:\\
"""\\
\{thought\_text\}\\
"""\\\\
Based on this text, determine which language the model decided to use for its response.\\
Classify into ONE of these categories:\\
\hspace*{1em} - English: if the model decided to respond in English\\
\hspace*{1em} - Korean: if the model decided to respond in Korean\\
\hspace*{1em} - Chinese: if the model decided to respond in Chinese\\
\hspace*{1em} - Others: if the model decided to respond in another language, or if the decision is unclear/mixed/not stated\\\\
Output your answer as a single JSON object with the format:\\
\{\{"language": "<English|Korean|Chinese|Others>", "confidence": "<high|medium|low>", "reason": "<brief explanation>"\}\}\\\\
Only output the JSON, nothing else.
\end{tcolorbox}

\subsection{Explicit Linguistic Instruction Prompts}
\label{app:zero_shot_few_shot}
We provide the system prompts used for zero-shot and few-shot explicit linguistic instruction experiments (Section~\ref{sec:explicit_linguistic_instruction}).

\begin{tcolorbox}[breakable, enhanced, top=1pt, left=1pt, right=1pt, bottom=1pt, title=Zero-shot System Prompt for Language Selection]
\small
You are a helpful multilingual AI assistant.

Your top priority is to choose the correct response language when the user's input contains code-switching (mixing two or more languages).
You must first determine the response language internally, and then produce the final answer only in that language.

Definitions
\begin{itemize}[leftmargin=*, topsep=2pt, itemsep=1pt]
    \item Code-switching: Using two or more languages within a single conversational context.
    \item Simple setting: Intra-sentential code-switching, where two languages are mixed within a single sentence or short utterance.
    \item Complex setting: The user query consists of two parts in different languages: an Instruction and Content.
\end{itemize}

Language Selection Policy

A. Simple setting (intra-sentential code-switching)
\begin{enumerate}[leftmargin=*, topsep=2pt, itemsep=1pt]
    \item Apply the Matrix Language Frame (MLF) principle to distinguish the Matrix Language (ML) and the Embedded Language (EL).
    \item Identify the Matrix Language (ML): the grammatically dominant language that provides the main syntactic structure (e.g., word order, function words, particles, tense markers, agreement, sentence-ending forms).
    \item Identify the Embedded Language (EL): inserted words or phrases from another language.
    \item Respond only in the Matrix Language.
\end{enumerate}

B. Complex setting (Instruction vs.\ Content)
\begin{enumerate}[leftmargin=*, topsep=2pt, itemsep=1pt]
    \item Segment the input into:
    \begin{itemize}[leftmargin=*, topsep=1pt, itemsep=1pt]
        \item Instruction: what the user asks the model to do (e.g., summarize, explain, rewrite, continue, fix grammar).
        \item Content: the text or material being discussed or manipulated.
    \end{itemize}
    \item Determine the task type and expected response language:
    \begin{itemize}[leftmargin=*, topsep=1pt, itemsep=1pt]
        \item Understanding / clarification tasks $\rightarrow$ respond in the Instruction Language.
        \item Direct manipulation / generation tasks $\rightarrow$ respond in the Content Language.
    \end{itemize}
\end{enumerate}

Output Constraints
\begin{itemize}[leftmargin=*, topsep=2pt, itemsep=1pt]
    \item Do not mention internal reasoning, language selection criteria, or decision process.
    \item Do not output the language decision explicitly.
    \item Do not mix languages in the response unless explicitly requested.
\end{itemize}
\end{tcolorbox}

\paragraph{Few-shot Demonstrations.}
We additionally prepend four task-agnostic demonstrations to the same system prompt in the few-shot setting.
All examples are disjoint from the benchmark and are used only to illustrate response language selection behavior.

\begin{tcolorbox}[breakable, enhanced, top=1pt, left=1pt, right=1pt, bottom=1pt,
title=Few-shot Demonstrations (Korean Matrix)]
\small
Example 1: \\
User: Machine learning에서 overfitting을 어떻게 방지할 수 있나요? \\
Assistant: 과적합을 방지하는 몇 가지 방법이 있습니다. 첫째, 정규화 기법(L1, L2)을 사용할 수 있습니다. 둘째, 드롭아웃을 적용하여 뉴런의 일부를 무작위로 비활성화합니다. 셋째, 교차 검증을 통해 모델을 평가합니다. 마지막으로, 더 많은 학습 데이터를 확보하는 것도 도움이 됩니다.

\vspace{0.5em}
Example 2: \\
User: Python으로 web scraping하려면 어떤 library를 써야 해? \\
Assistant: 웹 스크래핑을 위해 여러 라이브러리를 사용할 수 있습니다. BeautifulSoup은 HTML 파싱에 적합하고, Requests는 HTTP 요청을 보내는 데 사용됩니다. 동적 웹 페이지의 경우 Selenium이나 Playwright를 추천합니다.
\end{tcolorbox}

\begin{tcolorbox}[breakable, enhanced, top=1pt, left=1pt, right=1pt, bottom=1pt,
title=Few-shot Demonstrations (English Matrix)]
\small
Example 1: \\
User: Can you explain the concept of 애교? Why is it so popular among 케이팝 아이돌? \\
Assistant: Aegyo is a form of emotional expression commonly used by children or young people. It is particularly popular among K-pop idols because it helps enhance their charm and appeal.

\vspace{0.5em}
Example 2: \\
User: What should I wear to 장례식 and what are the 예절 for the 장례식 in the U.S.? \\
Assistant: When attending a funeral, it is customary to wear black or dark clothing. The family may provide specific instructions on dress code, so it's best to follow their guidance.
\end{tcolorbox}

\begin{tcolorbox}[breakable, enhanced, top=1pt, left=1pt, right=1pt, bottom=1pt,
title=Few-shot Demonstrations (Understanding Tasks)]
\small
Example 1 (Summarization -- respond in Instruction Language): \\
User: Could you give me a short summary of the following text?: \\
인공지능(AI)은 인간의 학습능력, 추론능력, 지각능력을 인공적으로 구현하려는 컴퓨터 과학의 한 분야이다. 현대에서 인공지능은 기계학습과 딥러닝의 발전으로 인해 급속한 발전을 이루었다. \\
Assistant: Artificial Intelligence (AI) is a branch of computer science that aims to artificially implement human learning, reasoning, and perception abilities. In modern times, AI has achieved rapid development due to advances in machine learning and deep learning.

\vspace{0.5em}
Example 2 (Explanation -- respond in Instruction Language): \\
User: 다음 내용을 설명해줘: \\
The greenhouse effect is a natural process that warms the Earth's surface. When the Sun's energy reaches the Earth's atmosphere, some of it is reflected back to space and some is absorbed. \\
Assistant: 온실 효과는 지구 표면을 따뜻하게 하는 자연적인 과정입니다. 태양 에너지가 지구 대기에 도달하면 일부는 우주로 반사되고 일부는 흡수됩니다. 이 과정을 통해 지구는 생명체가 살 수 있는 적정 온도를 유지합니다.
\end{tcolorbox}

\begin{tcolorbox}[breakable, enhanced, top=1pt, left=1pt, right=1pt, bottom=1pt,
title=Few-shot Demonstrations (Manipulation Tasks)]
\small
Example 1 (Continue writing -- respond in Content Language): \\
User: Continue the following story: \\
옛날 어느 작은 마을에 현명한 할아버지가 살고 있었습니다. 마을 사람들은 어려운 일이 있으면 항상 할아버지를 찾아갔습니다. \\
Assistant: 어느 날, 마을에 큰 가뭄이 들었습니다. 농작물이 말라가고 우물도 바닥을 드러내기 시작했습니다. 걱정에 휩싸인 마을 사람들은 할아버지를 찾아가 조언을 구했습니다. 할아버지는 잠시 생각에 잠기더니 ``산 너머 동굴에 숨겨진 샘이 있다''고 말씀하셨습니다.

\vspace{0.5em}
Example 2 (Rewrite -- respond in Content Language): \\
User: 주어진 내용을 더 공식적인 언어로 다시 작성해줘: \\
The data shows that users really like the new feature. They're using it a lot more than we expected. \\
Assistant: The data indicates that users have responded positively to the new feature. Usage metrics have exceeded initial projections significantly.
\end{tcolorbox}

\end{document}